\documentclass{ieeeaccess}
\usepackage{booktabs}
\usepackage{cite}
\usepackage{amsmath,amssymb,amsfonts}
\usepackage{algorithmic}
\usepackage{graphicx}
\usepackage{textcomp}
\usepackage{hyperref}
\usepackage[table]{xcolor}
\usepackage{subfig}
\usepackage{subcaption}
\usepackage{xcolor}
\usepackage{multirow}
\usepackage{balance}

\usepackage{bm}
\makeatletter
\AtBeginDocument{\DeclareMathVersion{bold}
	\SetSymbolFont{operators}{bold}{T1}{times}{b}{n}
	\SetSymbolFont{NewLetters}{bold}{T1}{times}{b}{it}
	\SetMathAlphabet{\mathrm}{bold}{T1}{times}{b}{n}
	\SetMathAlphabet{\mathit}{bold}{T1}{times}{b}{it}
	\SetMathAlphabet{\mathbf}{bold}{T1}{times}{b}{n}
	\SetMathAlphabet{\mathtt}{bold}{OT1}{pcr}{b}{n}
	\SetSymbolFont{symbols}{bold}{OMS}{cmsy}{b}{n}
	\renewcommand\boldmath{\@nomath\boldmath\mathversion{bold}}}
\makeatother

\def\BibTeX{{\rm B\kern-.05em{\sc i\kern-.025em b}\kern-.08em
		T\kern-.1667em\lower.7ex\hbox{E}\kern-.125emX}}

\begin{document}
	\history{Date of publication xxxx 00, 0000, date of current version xxxx 00, 0000.}
	\doi{1234567654321}
	
	\title{Seeing the Unseen: Towards Training-Free Inspection for Wind Turbine Blades Using Knowledge-Augmented Vision Language Models}
	\author{\uppercase{Yang Zhang},\IEEEmembership{Member, IEEE}, \uppercase{Qianyu Zhou}, \uppercase{Farhad Imani} and \uppercase{Jiong Tang}, \IEEEmembership{Member, IEEE}}
	
	\address[]{School of Mechanical, Aerospace and Manufacturing Engineering, University of Connecticut, Storrs, CT 06269, USA}
	
	\tfootnote{This research is supported in part by the U.S. Economic Development Administration through the Connecticut Manufacturing Simulation Center (CMSC) at the University of Connecticut, and in part by the National Science Foundation under grant 2434519.}
	
	\markboth
	{Zhang \headeretal: Preparation of Papers for IEEE TRANSACTIONS and JOURNALS}
	{Zhang \headeretal: Preparation of Papers for IEEE TRANSACTIONS and JOURNALS}
	
	\corresp{Corresponding author: Jiong Tang (e-mail: jiong.tang@uconn.edu).}

	\begin{abstract}
		Wind turbine blades operate in harsh environments, making timely damage detection essential for preventing failures and optimizing maintenance. Drone-based inspection and deep learning are promising, but typically depend on large, labeled datasets, which limit their ability to detect rare or evolving damage types. To address this, we propose a training-free, knowledge-grounded inspection framework that integrates Retrieval-Augmented Generation (RAG) with Vision-Language Models (VLM). A multimodal knowledge base is constructed, comprising technical documentation, representative reference images, and domain-specific guidelines. A hybrid text--image retriever with keyword-aware reranking assembles the most relevant context to condition the VLM at inference, injecting domain knowledge without task-specific training. We evaluate the framework on 110 labeled blade images covering four mechanism-based damage categories: healthy state, surface damage, environmental damage, and structural damage. The Full RAG framework achieves an overall accuracy of 94.55\%, with a macro-averaged precision of 0.9786, recall of 0.8750, and F1 score of 0.9055. Ablation studies demonstrate that both the hybrid text--image retrieval mechanism and the keyword-aware reranking step are individually necessary for robust multi-class damage identification, with single-modality retrieval leading to systematic failure on specific damage categories. Comparison against AnomalyCLIP confirms that structured domain knowledge integration is the key differentiating factor over general-purpose zero-shot vision models. Comparison against YOLO-based detectors further highlights the generalization limitations of supervised pipelines across imaging conditions and damage types. The proposed framework requires no task-specific model adaptation, supports on-the-fly knowledge base updates, and produces structured diagnostic reports with traceable reasoning, offering a data-efficient and interpretable solution for industrial blade inspection.
	\end{abstract}
	
	
	\titlepgskip=-21pt
	
	\maketitle
	
	\section{Introduction}
	\label{sec:introduction}
	
	\PARstart{W}{ind} energy has become a key component of the global transition toward sustainable power. With rapid growth in installed capacity and the increasing scale of wind farms, ensuring the reliability and durability of wind turbine components is more important than ever. Among these components, turbine blades are particularly vulnerable due to exposure to harsh environmental conditions, including ultraviolet radiation, rain, hail, lightning strikes, and temperature cycling. These stressors lead to diverse set of failure modes and damage patterns, such as leading-edge erosion, surface cracks, damaged lightning receptors, and delamination, among others. If undetected, such damage can compromise structural integrity, increase downtime, reduce energy output, and raise maintenance costs. Timely detection and assessment are therefore essential to maintain operational efficiency and safety \cite{b1,b2,b3}. A wide range of inspection and monitoring techniques has been explored for wind turbine blade evaluation, including acoustic emission sensing \cite{b4}, ultrasonic testing \cite{b5,b6}, and infrared thermography \cite{b7}. These traditional methods are reliable in many scenarios and are increasingly enhanced by artificial intelligence. In contrast, vision-based approaches offer complementary advantages through their non-contact operation, high spatial resolution, full-surface coverage, and potential for near real-time defect identification. Machine vision also offers enhanced interpretability, allowing visual outputs to be directly assessed by human operators or automated systems. Furthermore, the advances in drone technologies enable flexible, automated, and scalable image acquisition from multiple viewing angles without requiring turbine shutdown \cite{b8, b9, b10}. Combined with AI-powered image analysis, this approach provides a safe, efficient, and non-intrusive solution for large-scale blade inspection, aligned with the practical needs of modern wind operations.
	
	Recent advances in computer vision and deep learning have accelerated the development of automated blade inspection systems. Multiple architectures have shown promise, with object detection frameworks such as YOLOv8 widely adapted via specialized enhancements for damage detection \cite{b11,b12,b13, b135}. Multimodal approaches are also effective, with researchers integrating optical-thermal video fusion \cite{b14}, visible-infrared image fusion \cite{b15}, and hyperspectral imaging with 3D CNNs \cite{b16} to improve recognition under challenging field conditions. Attention-based models have emerged as powerful tools, with Vision Transformers outperforming traditional CNNs in surface defect classification \cite{b17} and various attention mechanisms being incorporated into existing architectures to enhance feature extraction \cite{b18}. Beyond supervised learning, unsupervised techniques, such as memory-aided denoising autoencoders \cite{b19} and reverse knowledge distillation \cite{b20}, show promise in limited labeled settings, demonstrating the field’s continued evolution toward robust and practical deployments. A comprehensive list of related studies is provided in Table ~\ref{tab:wind_turbine_inspection}. Despite this progress, effectiveness still depends heavily on access to well‑labeled, balanced, and high‑quality datasets, a condition rarely met in realistic wind turbine environments. In practice, collecting representative and consistent training data poses significant challenges. Factors such as environmental variability, seasonal shifts, inconsistent drone viewpoints, and inspection scheduling contribute to uneven and often noisy datasets. Certain defects (e.g., surface stains) occur frequently and are easily captured, whereas more critical damage types (e.g., cracks, delamination, or lightning-induced erosion) occur infrequently and are harder to document. The result is a persistent and dynamic data imbalance, which undermines the performance and generalization capability of conventional deep learning pipelines.
	
	\begin{table*}[htbp]
		\centering
		\caption{Wind turbine blade inspection using deep learning and vision techniques.}
		\label{tab:wind_turbine_inspection}
		\setlength{\tabcolsep}{3pt}
		\begin{tabular}{|p{60pt}|p{200pt}|p{200pt}|}
			\hline
			\textbf{Papers} & \textbf{Techniques} & \textbf{Applications} \\
			\hline
			\cite{b11,b12,b13,b18,b21,b22,b23} & YOLOv8, YOLOv7, YOLOv5 with SE attention, GSConv, EMA, GA optimization & Wind turbine blade damage detection \\
			\hline
			\cite{b15} & YOLOv7 with RGB-IR feature fusion & Multimodal wind turbine defect detection \\
			\hline
			\cite{b24} & YOLOv5s with semi-supervised learning & Blade defect detection with limited labeled data \\
			\hline
			\cite{b17} & Vision Transformers (ViT) & Surface defect detection in renewable energy assets \\
			\hline
			\cite{b14,b19,b25} & AQUADA-Seg, Memory-Aided Denoising Autoencoder with Swin Transformer U-Net, Siamese CNN with similarity learning & Blade segmentation and damage detection \\
			\hline
			\cite{b16} & 3D CNN with hyperspectral imaging & Fault detection (cracks, erosion, ice) \\
			\hline
			\cite{b28} & Coarse-to-fine stitching with regression-based shape optimization & Drone-based image stitching for defect analysis \\
			\hline
			\cite{b20} & ResNet architectures with reverse knowledge distillation & Structural anomaly detection and localization \\
			\hline
			\cite{b29} & RARNN (Receptive Attention Recurrent Neural Network) & Digital twin for dynamic impact identification \\
			\hline
			\cite{b30} & Spatio-temporal attention model & Ice formation detection on wind turbine blades \\
			\hline
		\end{tabular}
	\end{table*}

	The challenges above have prompted growing interest in zero-shot approaches for blade inspection and health monitoring. Recent vision-language models, including AnomalyCLIP \cite{b31} and FiLo \cite{b32}, demonstrate promising capabilities. AnomalyCLIP leverages object-agnostic text prompts to detect anomalies across diverse domains, while FiLo incorporates large language models (LLMs) to provide fine-grained descriptions with enhanced localization. Similarly, GAN-based zero-shot transfer learning has shown strong performance for structural health monitoring, where researchers reported F1 scores of 0.978 \cite{b33} through domain adaptation and spectral mapping. More recent work has explored multi-source transfer learning \cite{b34} and autoencoder-based domain adaptation frameworks \cite{b35}, with some approaches achieving high accuracy even with unseen damage classes \cite{b36}. All zero-shot related anomaly or damage detection studies are listed in Table ~\ref{tab:anomaly_zeroshot}. However, these methods have limitations. GANs and similar generative models often demand extensive computational resources and still require a minimum amount of real data to generate realistic outputs. Zero-shot vision-language models may struggle to generalize to industrial domains underrepresented in pretraining data, and their outputs can lack grounding in domain-specific context. In practice, these limitations constrain scalability, robustness, and interpretability, especially in data-scarce, variable, or evolving inspection environments. There is therefore a pressing need for frameworks that make effective use of limited labeled data and expert knowledge, while remaining adaptable without retraining.
	
	\begin{table*}[htbp]
		\centering
		\caption{Anomaly detection using zero-shot learning.}
		\label{tab:anomaly_zeroshot}
		\setlength{\tabcolsep}{3pt}
		\begin{tabular}{|p{60pt}|p{200pt}|p{200pt}|}
			\hline
			\textbf{Papers} & \textbf{Techniques} & \textbf{Applications} \\
			\hline
			\cite{b31} & Vision-language model fine-tuning with object-agnostic prompt optimization & Zero-shot anomaly detection \\
			\hline
			\cite{b32} & Vision-language model training with Grounding DINO & Zero-shot anomaly detection with fine-grained descriptions \\
			\hline
			\cite{b33,b34} & GAN-based data generation with feature alignment & Zero-shot structural damage detection \\
			\hline
			\cite{b35} & Zero-shot CNN with domain adaptation & Cross-domain damage diagnosis \\
			\hline
			\cite{b36} & Generalized zero-shot learning (GZSL) with CNN backbones (ResNet, VGG, DenseNet) & Structural damage assessment with unseen classes \\
			\hline
			\cite{b37} & Image-text alignment with LVLM inference & Zero-shot industrial anomaly detection \\
			\hline
		\end{tabular}
	\end{table*}
	
	Recently, large language models have emerged as a compelling alternative, offering strong generalization capabilities with minimal supervision. LLMs are pretrained on massive corpora and can perform a wide range of tasks using only natural language prompts, often without additional fine‑tuning. These tasks include question answering, summarization, and anomaly detection, often without requiring additional fine-tuning. This flexibility makes LLMs attractive for rapid deployment in domains with limited labeled data. However, a fundamental limitation remains: LLMs are trained on broad, general-purpose data, and often lack the domain-specific grounding required for high-stakes applications such as structural health monitoring or wind turbine inspection. This gap can lead to factual inaccuracies or so-called hallucinations in industrial applications \cite{b38}, where models generate fluent but incorrect or unsupported outputs. To address these limitations, Retrieval-Augmented Generation (RAG) has emerged as a promising framework. By augmenting LLMs with external knowledge retrieved at inference time, RAG enables the model to ground its outputs in task-specific information without modifying its underlying parameters. This approach has been applied successfully in several recent studies. For instance, SafeLLM \cite{b39} introduces a domain-specific safety monitoring framework for offshore wind maintenance. It leverages LLMs together with statistical techniques to identify potentially unsafe or hallucinated responses. Pastoriza et al \cite{b40} developed a retrieval-augmented anomaly detection system, which incorporates human-in-the-loop feedback for continuous error correction. Similarly, Thimonier et al \cite{b41} applied retrieval-augmented learning to deep anomaly detection in tabular data, using transformer-based reconstructions grounded in retrieved context. Other promising approaches include AnomalyGPT \cite{b37}, which leverages large vision-language models for industrial anomaly detection and reports 86.1\% accuracy on benchmark datasets. Another example is LLM-DSKB \cite{b42}, which integrates LLMs with domain-specific knowledge bases for industrial equipment operation and maintenance. Liu et al \cite{b425} integrated knowledge graphs with LLMs for fault diagnosis in aviation assembly, demonstrating that retrieval-augmented reasoning over domain-specific structured knowledge significantly improves diagnostic accuracy in complex industrial settings. A summary regarding LLM and RAG in industrial applications is listed in Table ~\ref{tab:rag_llm}. 
	
	\begin{table*}[htbp]
		\centering
		\caption{RAG and LLM in industrial applications.}
		\label{tab:rag_llm}
		\setlength{\tabcolsep}{3pt}
		\begin{tabular}{|p{60pt}|p{200pt}|p{200pt}|}
			\hline
			\textbf{Papers} & \textbf{Techniques} & \textbf{Applications} \\
			\hline
			\cite{b39,b43} & Statistical safety measures with Wasserstein distance and cosine similarity using Universal Sentence Encoder & Domain-specific safety monitoring for offshore wind maintenance \\
			\hline
			\cite{b40} & Retrieval-augmented post-processing & Anomaly detection adjustment \\
			\hline
			\cite{b41} & Transformer-based anomaly reconstruction with retrieval-enhanced scoring & Anomaly detection in structured tabular data \\
			\hline
			\cite{b42} & LLM embeddings with vector retrieval & Domain-adapted industrial equipment maintenance \\
			\hline
			\cite{b44} & Time-series to text conversion with LLM prompting and forecasting & Zero-shot time series anomaly detection \\
			\hline
			\cite{b45} & Multimodal LLM approach for contextual understanding and information extraction & Fault detection and diagnostics in hydrogenator \\
			\hline
		\end{tabular}
	\end{table*}

	In addition to these industrial applications, recent studies have investigated RAG-grounded VLMs in broader visual tasks. For example, Visual RAG \cite{b46} demonstrates how multimodal large models can expand visual knowledge without fine-tuning by retrieving relevant exemplars at inference time. Bhat et al \cite{b47} integrated RAG with VLMs for scientific visual question answering, significantly improving factual grounding. Similarly, Dong et al \cite{b48} introduced semantic document layout analysis to enhance visually rich RAG tasks, while Khan et al \cite{b49} applied retrieval-augmented multimodal reasoning to open-vocabulary species recognition, achieving notable gains on unseen categories. These developments underscore the growing interest in RAG-grounded VLMs for visual understanding tasks \cite{b50}. Wind turbine blades present unique inspection challenges that differ from conventional anomaly detection tasks. Their large structural scale, diverse damage modalities (e.g., cracks, corrosion, peeling, and composite delamination), and highly variable environmental conditions complicate the design of robust detection systems. Traditional supervised learning approaches struggle to keep pace with these evolving and heterogeneous failure patterns, because they require continuous data collection and retraining. Vision-language models, when augmented with retrieval mechanisms, are well positioned to address these challenges by linking visual semantics from inspection imagery with structured domain knowledge, thereby enabling more adaptive, interpretable, and generalizable inspection capabilities. Nevertheless, despite their potential, the integration of RAG-grounded VLMs into the wind energy sector remains largely unexplored. 
	
	To fill this gap, we propose a multimodal retrieval-grounded visual reasoning framework for wind turbine blade inspection, enabling interpretable training-free diagnosis. Unlike conventional text-only RAG, our system integrates textual and visual knowledge within a structured hybrid knowledge base that includes damage descriptions, turbine information, maintenance guidelines, and annotated exemplars. Knowledge entries are embedded with dual encoders (text and image) and stored in a vector database for efficient cross-modal retrieval. A domain-aware reranking mechanism further refines the results to ensure each inspection is guided by the most relevant evidence. The retrieved context is incorporated into a dynamic prompt that balances general expertise with visually similar reference cases, guiding a vision-language model to produce structured diagnostic reports covering blade count, damage presence, type, severity, and explanatory rationale. The architecture requires no task-specific retraining and supports on-the-fly updates through simple additions to the knowledge base. Importantly, the system preserves transparency by tracing which knowledge items informed each decision, creating a clear provenance from evidence to conclusion. In experiments, the framework demonstrates clear advantages over both supervised detectors and open-vocabulary vision models, particularly for rare or complex damage categories where labeled data are scarce. Although our evaluation set is necessarily limited due to the difficulty of obtaining verified blade images, it spans diverse defect types, and results are supported by baseline comparisons and uncertainty analysis. Together, these elements highlight the promise of RAG-grounded VLMs for creating adaptive, data-efficient, and interpretable inspection systems tailored to the unique challenges of the wind energy sector.
	
	The remainder of this paper is organized as follows. Section 2 presents the overall architecture of the proposed visual-text RAG framework, including the design of the hybrid knowledge base, embedding strategy, retrieval mechanism, and prompt construction. Section 3 details the experimental setup, including dataset preparation, baseline models, and implementation specifics. Section 4 reports and discusses the evaluation results, highlighting the effectiveness of the proposed method under limited data conditions. Finally, Section 5 concludes the paper and outlines future directions for expanding domain adaptability and real-world deployment.

	\section{Framework of Wind Turbine Blade Inspection with RAG-Grounded VLM}
	\subsection{Overall architecture}
\begin{figure*}[htbp]
		\centering
		\includegraphics[width=\textwidth]{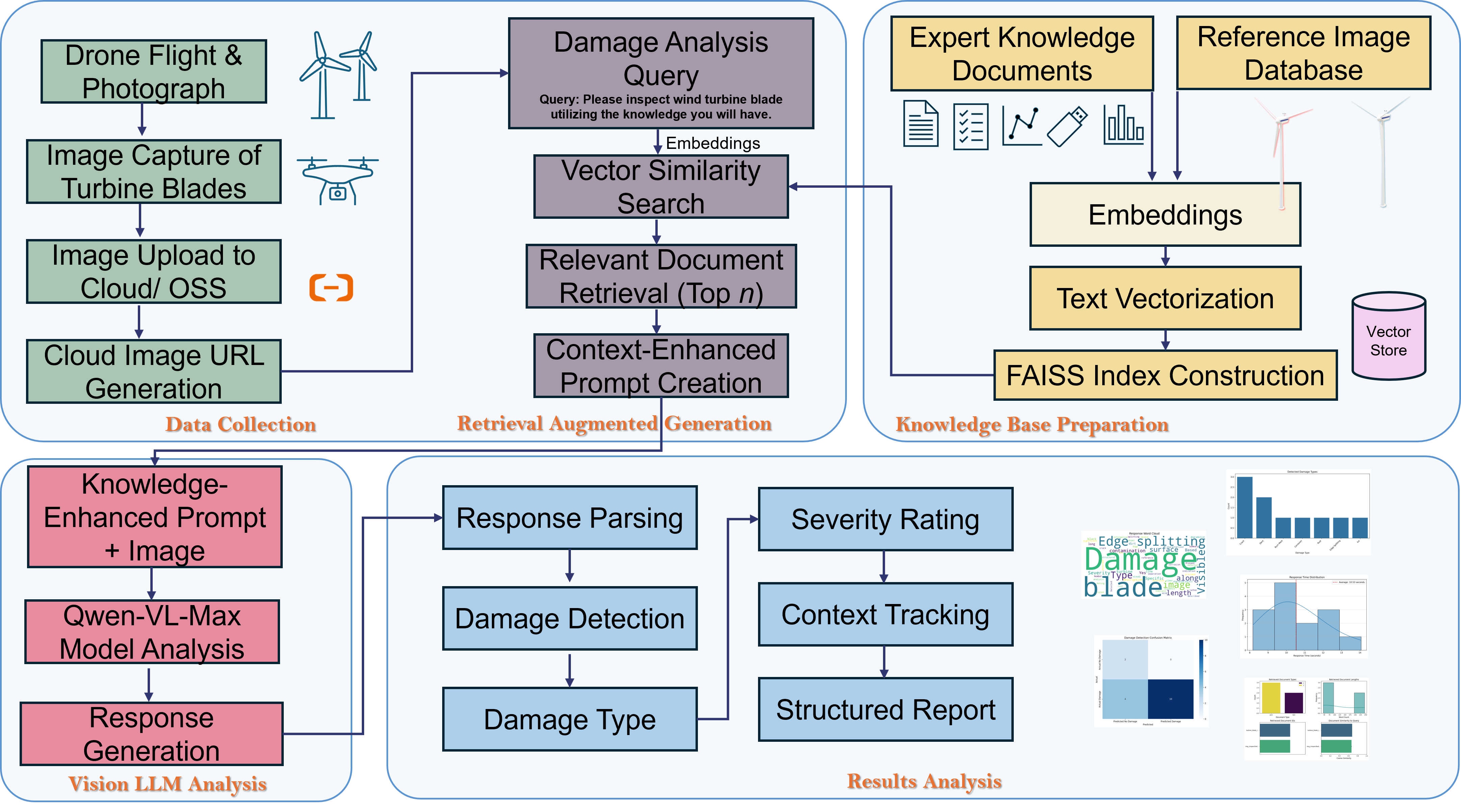}
		\caption{General flowchart for wind turbine blade inspection using RAG and VLM.}
		\label{fig:f1}
	\end{figure*}

	Traditional inspection methods struggle with the complex nature of wind turbine blade damage detection under diverse environmental conditions. To address these challenges, we design a multimodal RAG-grounded VLM framework that integrates textual and visual knowledge for accurate damage assessment. As shown in Figure ~\ref{fig:f1}, the architecture follows an end-to-end pipeline with four main stages: (1) Data Collection, where drone-captured blade images are uploaded and prepared for analysis; (2) Knowledge Base Preparation, in which expert documents and annotated reference images are encoded and indexed for efficient retrieval; (3) Retrieval-Augmented Inference, where damage queries trigger cross-modal similarity search and domain-aware reranking to assemble the most relevant evidence; and (4) Vision-Language Reasoning, where the enriched context and input image are processed by a VLM to generate structured diagnostic reports, including damage detection, type, severity, and descriptive explanation. This pipeline enables training-free assessment by grounding visual analysis in domain knowledge.

	\subsection{Retrieval augmented generation}
\begin{figure*}[htbp]
		\centering
		\includegraphics[width=\textwidth]{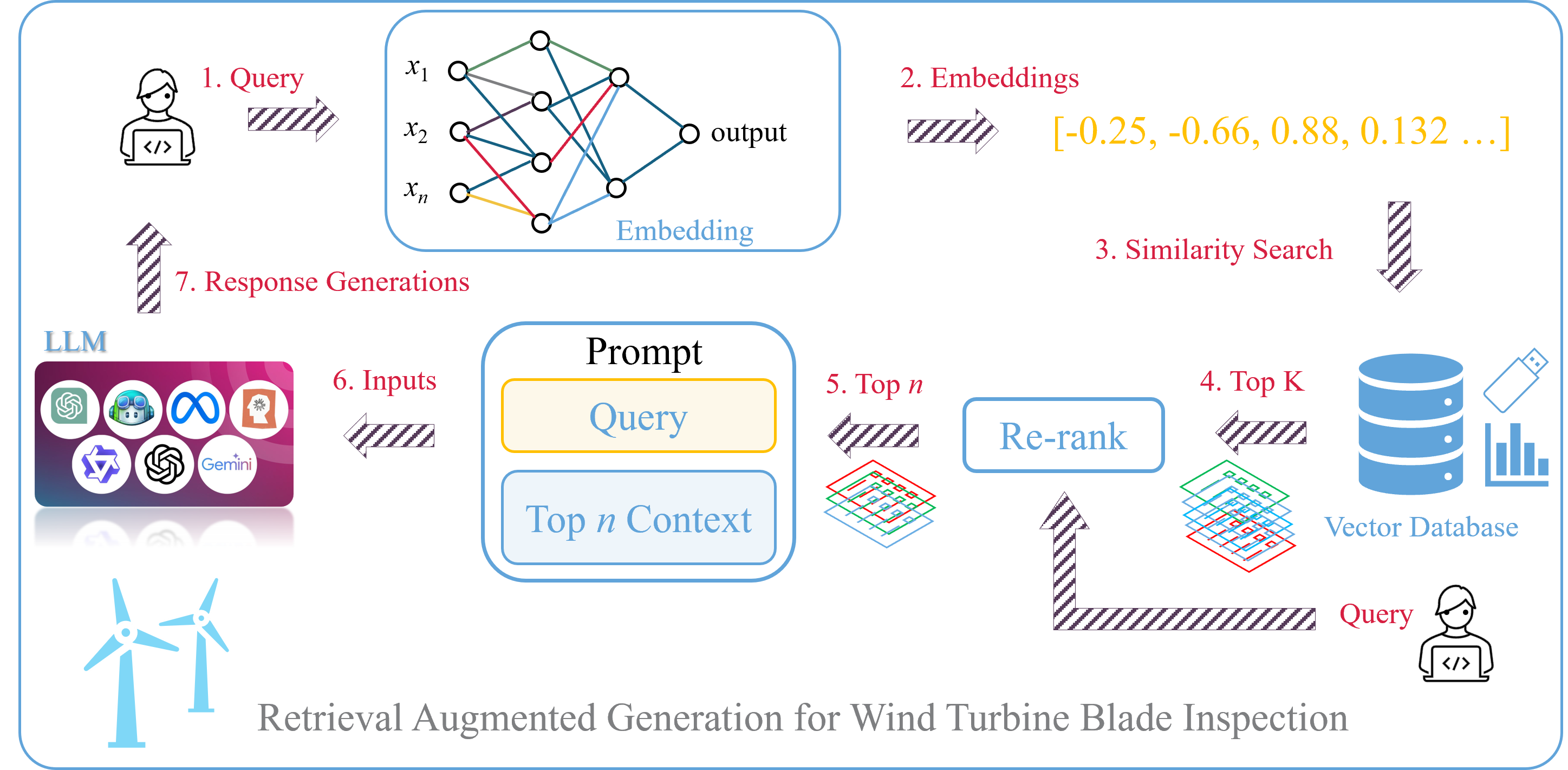}
		\caption{Schematic diagram of retrieval augmented generation.}
		\label{fig:f2}
	\end{figure*}

	RAG is the foundation of our framework, allowing the vision-language model to ground its analysis in domain-specific knowledge that pre-trained models alone cannot provide. By retrieving relevant information at inference time, RAG enables accurate assessment without task-specific fine-tuning. The implementation extends conventional RAG by incorporating multimodal retrieval, combining textual descriptions (e.g., classification criteria) with visual exemplars (e.g., images of damage types) to provide complementary evidence for blade inspection. The RAG pipeline comprises three components: (1) a structured knowledge base with domain-specific documentation and reference images, (2) a vector database for efficient embedding-based retrieval, and (3) a similarity search with reranking to identify the most relevant context for each query. Figure ~\ref{fig:f2} shows the schematic flow of this system, and the following subsections describe each component in detail.
	
	\subsubsection{Knowledge base} 
	The knowledge base serves as the repository of domain-specific information that the system can access during analysis. We designed a dual-modal knowledge base that incorporates both textual and visual information relevant to wind turbine blade damage assessment. The textual component of the knowledge base contains several types of information. Lists of contents from knowledge base are provided in Table ~\ref{tab:knowledge_base}. There are totally 4 knowledge bases, with each functioning in different roles. For example, the damage description knowledge base (Textual 1) stores the descriptions of commonly seen damage types in wind turbine blades. The descriptions detail how the damage should look like, what the color of that damage will be, and what the shape of that damage type is, etc. Additionally, we provide the basic information of the wind turbine itself (Textual 2), such as what material the blade is made of, what the blade looks like under healthy condition (painted white), etc. Therefore, when VLM fetches this knowledge, it can help the VLM distinguish the damage (which usually causes color change) from the healthy state. Furthermore, we provide maintenance logs (Textual 3), which record past events. In each event, the identified damage is labeled with a severity level. Although this is not an essential feature, it is often useful to obtain a rough estimate of damage severity, which can assist in maintenance planning and logistics. We also incorporate an image-textual database (Image-Textual Metadata) in this study. Since the datasets we use involve images that are taken in different visual conditions, such as sunny daytime, cloudy daytime, nighttime, and dusk, this multimodal knowledge base is essential for robust performance across varying lighting conditions. The image-textual database provides concrete visual examples that help the model recognize damage patterns despite visual variations due to weather, time of day, or camera position. This point is critical, as certain damage types exhibit different visual characteristics under varying lighting conditions, and textual descriptions alone are often inadequate for accurate identification. The visual component of the knowledge base consists of reference images showing various types of wind turbine blade damage. Each image is associated with metadata including {\bf Description}: A textual explanation of the specific damage features visible in the image; {\bf Image path}: A reference to the stored image file. In embeddings, the image will be fetched through the path. This metadata is stored in JSON format, enabling seamless integration with the textual knowledge base while maintaining the relationships between images and their corresponding damage descriptions.
	
	\begin{table*}[htbp]
		\centering
		\caption{Lists of expert knowledge base about wind turbine blades.}
		\label{tab:knowledge_base}
		\setlength{\tabcolsep}{3pt}
		\begin{tabular}{|p{120pt}|p{340pt}|}
			\hline
			\rowcolor{orange!30}
			\textbf{Knowledge Types} & \textbf{Descriptions (Dynamic)} \\
			\hline
			Image-Textual Metadata & $\bullet$ Images with known damage. The texts describe the images from different perspectives. \\
			\hline
			Textual 1 & $\bullet$ Texts describing damage types: color, shape, location etc. \\
			\hline
			Textual 2 & $\bullet$ Texts describing features of wind turbine, material, color, vortex generator, etc. \\
			\hline
			Textual 3 & $\bullet$ Texts describing maintenance logs and damage severity levels. \\
			\hline
		\end{tabular}
	\end{table*}

	To enhance retrieval efficiency, the textual documents are processed using a chunking mechanism based on the \textit{ RecursiveCharacterTextSplitter} from the LangChain library \cite{b51}, which is specifically designed for document processing in retrieval-augmented generation systems. The \textit {RecursiveCharacterTextSplitter} implements an intelligent recursive splitting algorithm that attempts to maintain the semantic integrity of content through hierarchical boundary detection. The \textit{ RecursiveCharacterTextSplitter} operates by attempting to split text at the most semantically meaningful boundaries first, working through the provided separators list in order. In implementation of this study, it first attempts to split at paragraph breaks (\textbackslash n\textbackslash n), then at line breaks (\textbackslash n), then at sentence boundaries (.), and finally at word boundaries ( ). Only if no suitable boundaries are found will it resort to character-level splitting (""). This hierarchical approach preserves the natural structure of the text as much as possible, which is crucial for maintaining the contextual meaning of technical documentation. This approach divides documents into manageable segments while maintaining semantic coherence by preferentially splitting at natural boundaries such as paragraph breaks. The chunk size parameter (set to 1000 characters) and chunk overlap parameter (set to 200 characters) are selected to balance retrieval precision with computational efficiency. The overlap between adjacent chunks ensures that concepts spanning chunk boundaries are not lost during retrieval. The knowledge base is designed to be extensible (as marked as Dynamic in Table ~\ref{tab:knowledge_base}), allowing new documents and reference images to be added through the adding document and adding reference image methods, respectively. This extensibility is critical for real-world deployment, as it enables the system to continuously incorporate new knowledge and examples as they become available, improving performance over time without requiring model retraining.
	
	\subsubsection{Embeddings and vector base}
	To enable efficient retrieval of relevant information from the knowledge base, we implemented a dual-embedding approach that captures both textual and visual semantic relationships. This section details the embedding models selected for each modality and the vector storage solution used for similarity search.
	
	\textit {Text embeddings}
	
	For encoding textual information, we select the Sentence-BERT all-MiniLM-L6-v2 model \cite{b52}. This model is chosen because it achieves a good balance between computational efficiency and embedding quality, making it suitable for resource-constrained industrial environments. It also performs well in capturing semantic relationships between texts, which is important for retrieving conceptually related information even when keywords do not match exactly. In addition, although the model is pre-trained on general text, it has demonstrated strong transferability to technical domains without requiring task-specific fine-tuning. The text embedding process then transforms each document chunk into a dense vector representation. Each document chunk is encoded into a fixed-dimensional vector (384 dimensions, referring to the dimensionality of the vector representation created by the model) that captures its semantic content. These vectors enable the system to identify conceptually similar documents even when they use different terminology, addressing a key limitation of traditional keyword-based retrieval systems.
	
	\textit {Image embeddings}
	
	For encoding visual information, we employ the CLIP (Contrastive Language-Image Pre-training) model \cite{b53}, specifically the openai/clip-vit-base-patch32 variant. CLIP is well suited to our framework because it learns a joint representation of text and images, which enables effective cross-modal retrieval. The Vision Transformer (ViT) architecture underlying CLIP captures rich visual features that are highly relevant for damage detection. Moreover, CLIP’s training strategy provides strong zero-shot generalization, which is crucial for recognizing rare or unusual damage patterns. Following extraction, a normalization procedure is applied to ensure consistency in similarity measurements during the retrieval process. The resulting embeddings are 512-dimensional vectors that capture the visual characteristics of each reference image. The choice of 512 dimensions for visual embeddings complements our text embedding dimensionality (384) while providing sufficient capacity to represent complex visual features. The higher dimensionality of visual data reflects its intrinsic complexity relative to textual content, enabling richer representation of spatial patterns, textures, and other fine-grained characteristics critical for effective damage analysis.
	
	\textit {Vector storage with FAISS}
	
	To enable efficient similarity search, we implemented vector storage using the FAISS library \cite{b54}. FAISS is selected for its computational efficiency, scalability and customization options. Specifically, FAISS implements optimized algorithms for high-dimensional similarity search, enabling rapid retrieval even with large knowledge bases. In addition, the library supports both in-memory and disk-based indices, allowing the system to scale to large collections of documents and images. Furthermore, FAISS offers various index types optimized for different retrieval scenarios, and in our implementation, we select the basic IndexFlatL2, which performs exact nearest neighbor search using the L2 (Euclidean) distance metric, due to its precision and the moderate size of our knowledge base.
	
	We create separate indices for textual and visual embeddings. Specifically, we select the basic IndexFlatL2 structure due to its precision and the moderate size of our knowledge base. The text index is configured to handle the 384-dimensional text embeddings, while the image index is designed specifically for the 512-dimensional visual embeddings. Both indices store their respective embeddings as float32 arrays, enabling efficient similarity searches across both modalities. The L2 distance metric is chosen for similarity calculations as it provides intuitive distance measurements in the embedding space and is compatible with the normalized embeddings produced by the models. The mathematical expression for L2 is is $L_2(\mathbf{x}, \mathbf{y}) = \sqrt{\sum_{i=1}^{n}(x_i - y_i)^2}$. Here $\mathbf{x}$ and $\mathbf{y}$ are the vectors being compared, with $x_i$ and $y_i$ representing their respective components at position $i$. This dual-embedding approach, combined with efficient vector storage, forms the foundation of our retrieval system, enabling the integration of diverse information sources during the damage assessment process.
	
	\subsubsection{Similarity search and rerank}
	The effectiveness of an RAG system is fundamentally determined by its ability to retrieve the most relevant information from the knowledge base. The approach proposed implements a two-stage retrieval process consisting of an initial similarity search followed by a reranking step that further refines the results.
	
	\textit{Hybrid similarity search}
	
	To leverage both textual and visual information during retrieval, we implement a hybrid similarity search mechanism that operates across both modalities simultaneously. This parallel retrieval approach is a key point in our system, as it allows for the integration of complementary information types during analysis. The similarity search process begins with a textual query formulation. For damage assessment tasks, we define a default query that captures the essential information needs: "\textit{comprehensive wind turbine blade damage assessment guidelines including technical documentation. The image to be analyzed may be taken at cloudy, night or dusk with bad vision}." This query is intentionally designed to retrieve broad contextual information about damage assessment while also accounting for challenging imaging conditions that are common in real-world wind turbine inspections. The query is encoded using the same text embedding model used for the knowledge base, transforming it into a 384-dimensional vector representation. This embedding is then used to perform a k-nearest neighbors search in the text index, retrieving the most semantically similar documents from our knowledge base using $\text{sim}(\mathbf{q}, \mathbf{d}) = \mathbf{q} \cdot \mathbf{d} / ||\mathbf{q}|| \cdot ||\mathbf{d}||$. $\mathbf{q}$ represents the query vector (the search input) and $\mathbf{d}$ represents the document vector (i.e., items in the knowledge base). In parallel, if an input image is provided, it is processed using the CLIP model to create a 512-dimensional visual embedding. This embedding is used to perform a similar k-nearest neighbors search in the image index, identifying visually similar reference images from our database. The result of these parallel searches is a set of potentially relevant text documents and a set of visually similar reference images. The parameter top\_$k$ controls the initial number of results retrieved from each modality, providing a balance between recall (retrieving all relevant items) and the computational cost of subsequent processing, as shown in Equations (1) and (2).
	\begin{align}
		R_{\text{text}} &= \text{topK}(\text{sim}(\mathbf{q}_{\text{text}}, \mathbf{d}_i)) \quad \forall \mathbf{d}_i \in D_{\text{text}} \\
		R_{\text{image}} &= \text{topK}(\text{sim}(\mathbf{q}_{\text{image}}, \mathbf{d}_j)) \quad \forall \mathbf{d}_j \in D_{\text{image}}
	\end{align}
	
	We set topK to 5 in our implementation to strike an optimal balance between retrieval comprehensiveness and computational efficiency, ensuring the system captures sufficient contextual information while maintaining responsiveness for real-time damage assessment applications.
	
	\textit{Reranking algorithm}
	
	While embedding-based similarity search is effective at identifying broadly relevant information, it may not optimally prioritize the most useful documents for a specific task. To address this, we implement a reranking algorithm that refines the initial retrieval results. The reranking approach combines multiple signals to assess the relevance of each retrieved document. Specifically, the algorithm considers two main factors. First, documents containing more query keywords are assigned higher scores, which helps prioritize the most topically relevant information. Second, shorter, and more focused documents are given slightly higher priority, as they often contain concentrated relevant content compared with longer, more general documents. While the initial retrieval phase operates in the vector space where documents and queries are represented as embeddings, the reranking phase works directly with the retrieved document objects and their textual content. In this second phase, we process the actual text rather than vector representations, allowing for content-based heuristics as shown in the following equations,
	\begin{equation}
		\text{keyword\_score}(d) = \sum_{k \in \text{keywords}} \text{Ind}(k \in \text{content}(d))
	\end{equation}
	
	\begin{equation}
		\text{length\_factor}(d) = \frac{1}{0.1 + \frac{|\text{content}(d)|}{1000}}
	\end{equation}
	
	\begin{equation}
		\text{score}(d, q) = \text{keyword\_score}(d) \cdot \text{length\_factor}(d)
	\end{equation}
	where Ind is an indicator function that returns 1 when keyword k appears in the document content and 0 otherwise; |content(d)|  represents the document length; $q$ denotes the query keyword set, and $d$ represents the document object. The division by 1000 normalizes document length to a practical scale, making the formula work effectively across documents of varying sizes. This approach is computationally efficient while still providing meaningful improvements over the initial embedding-based retrieval. The parameter top\_$n$ (set to 3 in the implementation) controls the final number of documents retained after reranking, focusing the context on the most relevant information using score obtained as show in Equation (6). 
	\begin{equation}
		R_{\text{final}} = \text{top\_}n(\text{score}(d, q)) \quad \forall d \in R_{\text{initial}}
	\end{equation}
	
	The combination of hybrid similarity search with reranking enables our system to efficiently identify the most relevant textual and visual information for a given damage assessment task, providing a rich contextual foundation for the analysis of vision-language model.
	
	\subsection{Response generation and result extractions}
	The final component of our system is the response generation module, which leverages a vision-language model to analyze the input image in conjunction with the retrieved context and produce a structured damage assessment. This section details the prompt engineering approach, the VLM integration, and the result extraction technique.
	
	\textit{Dynamic prompt construction}
	
	A critical aspect of our approach is the construction of effective prompts that guide the analysis of VLM. Rather than using static prompts, we adopt a dynamic prompt construction technique that incorporates retrieved contextual information. This approach enables the model to benefit from domain-specific knowledge while maintaining the flexibility to address diverse damage assessment scenarios. The prompt construction process starts with an initial base instruction: \textit{I need to utilize the knowledge base and observe the features of the anomaly on the wind turbine related components, and identify damage type.} This foundation establishes the core analytical objective. To achieve dynamic prompting, we append a transitional phrase: \textit{Using the following reference information to help with the analysis:} In this process, two key elements are incorporated: relevant textual knowledge and similar reference images. Textual knowledge is derived from the top-ranked documents, where the most relevant damage type information is extracted and formatted. Metadata from similar reference images, when available, is also integrated to provide comparative examples. This dynamic composition ensures each prompt is uniquely tailored to the specific damage assessment task while maintaining a consistent framework that guides the VLM analysis process. By combining task-specific instructions with contextually relevant knowledge, the proposed approach enables more accurate and informed damage assessments across diverse scenarios. The prompt including specific analysis instructions is shown in Table ~\ref{tab:dynamic_prompts}:
	
	\begin{table}[htbp]
		\centering
		\caption{Dynamic prompt constructions.}
		\label{tab:dynamic_prompts}
		\setlength{\tabcolsep}{3pt}
		\begin{tabular}{|p{240pt}|}
			\hline
			\rowcolor{orange!30}
			\textbf{Analytical Prompts} \\
			\hline
			"Based on these descriptions and references, analyze the image and determine:
			
			1. How many blades are visible in the image?
			
			2. Is there visible damage on any of the turbine blades in the image?
			
			3. If yes, what specific type of damage can be identified in this damage lists ('Missing Teeth of Vortex generators', 'Lightning Receptors', 'Crack', 'Corrosion', 'Erosion', 'Rust', 'Delamination', 'Fracture', 'Dent', 'Ice', 'Snow', 'Surface Peeling', 'Wear', 'Lightning Strike/Burning')?
			
			4. Provide a detailed description of the damage observed, referencing the specific characteristics described above.
			
			5. Rate the severity of the damage on a scale of 1-5, where 1 is minor and 5 is severe." \\
			\hline
		\end{tabular}
	\end{table}
	
	These structured questions serve multiple purposes. They provide a clear analytical framework that guides the assessment of VLM and ensure comprehensive coverage of key damage characteristics. In addition, they facilitate subsequent extraction of structured information from the responses generated by the model. When visually similar reference images are available, the prompt further includes adaptive guidance based on patterns in those references. This adaptive component enhances the system's intelligence in two ways. When all similar reference images show the same damage type, a targeted note is added to suggest that the VLM carefully check for this specific type. When the reference images display multiple damage types, these are presented as potential candidates, guiding the VLM to assess which, if any, are present.
	
	\textit{Result extraction and structuring}
	
\begin{figure*}[htbp]
		\centering
		\includegraphics[width=0.8\textwidth]{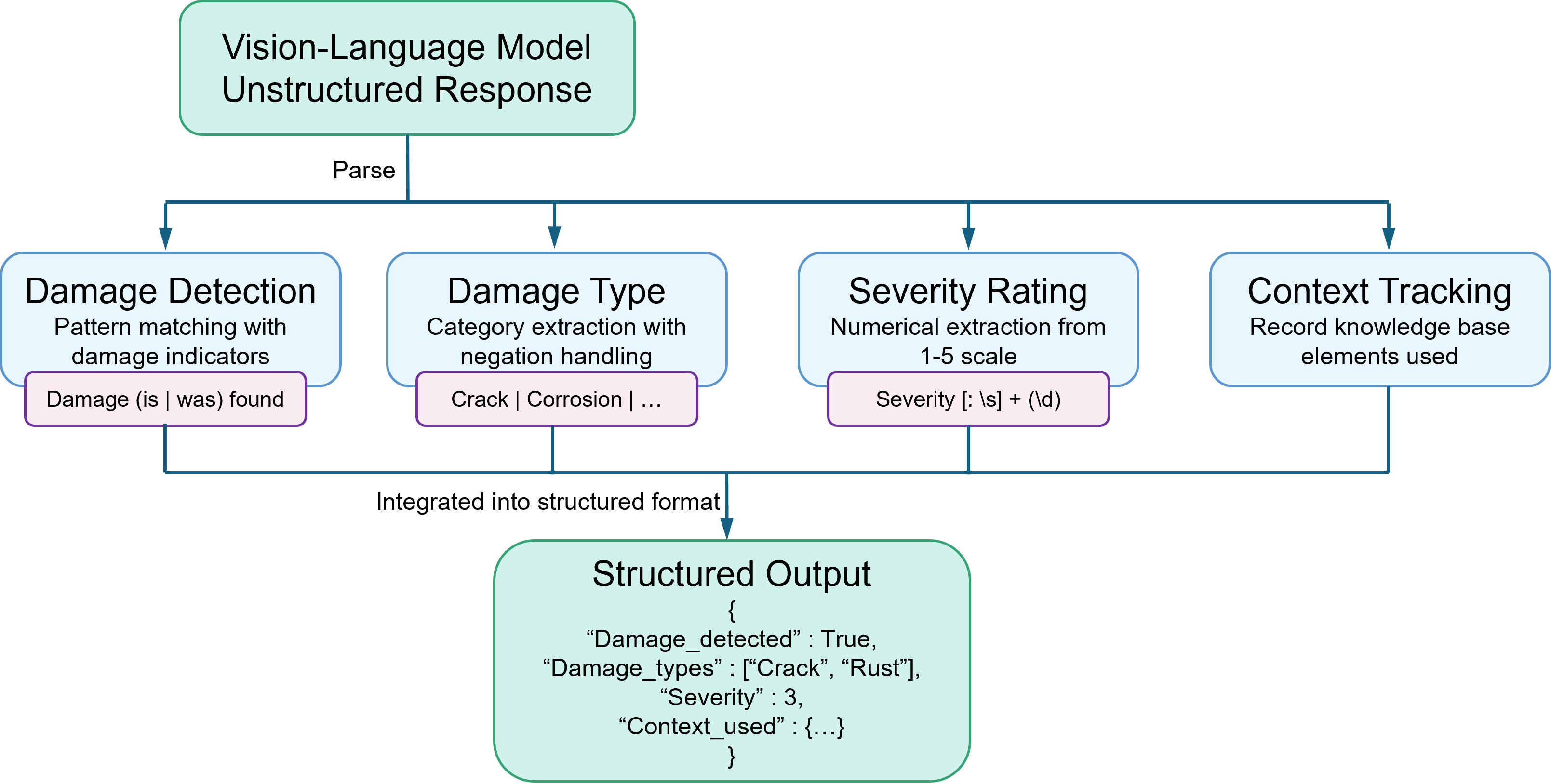}
		\caption{Structured output extraction process.}
		\label{fig:f3}
	\end{figure*}

	Our system integrates with the Qwen-VL-Max \cite{b55} vision-language model through an OpenAI-compatible API interface. The implementation passes both the constructed prompt and the image URL to the model, enabling comprehensive analysis that considers both visual evidence and contextual information. The image is fetched from cloud storage where it is pre-uploaded. This mimics the engineering implementation loop where the drone captures images and transmits them to the terminal, where the system then analyzes the visual data. Since the output from VLM is natural language, we implement a systematic extraction approach that transforms the free-text response into a structured format containing key assessment metrics. The structured output captures six essential elements: the complete model response, a damage detection flag, identified damage types, severity rating, descriptive assessment, and a record of knowledge base elements used during analysis. A schematic diagram is given to show the structured answer extraction process in Figure ~\ref{fig:f3}. 
	
	The extraction process employs natural language processing techniques to identify meaningful patterns in the model response. For damage detection, we analyze the text for indicative phrases such as "damage is detected," "there is damage," or "signs of damage." This approach provides a Boolean indicator of whether damage is detected in the image. For damage type classification, we implement a comprehensive pattern recognition system that searches for mentions of specific damage categories from our taxonomy, including cracks, corrosion, delamination, and others. Importantly, our system incorporates negation handling to prevent false positives. For example, phrases like "no cracks" or "absence of corrosion" are correctly interpreted as the absence of those damage types rather than their presence. Severity assessment is extracted through targeted pattern matching that identifies numerical ratings within context. The system searches for phrases like “severity: 3” or “severity rating of 4” to establish a quantified assessment on our predefined 1-5 scale.
	
	This extraction method is designed to be robust against variations in the model response format. By carefully considering contextual cues and linguistic constructions, we ensure accurate capture of assessment metrics regardless of the specific phrasing used by the model. The inclusion of negation handling is particularly important for technical assessments, as it prevents misinterpretation when the model explicitly notes the absence of damage characteristics. The resulting structured output significantly enhances the system integration capabilities with downstream applications such as maintenance management systems, inspection databases, or automated reporting tools. Furthermore, by tracking which knowledge base elements influenced the assessment through the "context\_used" field, we provide transparency that supports explainability and audit capabilities. This balanced approach to information extraction retains the model’s nuanced analysis and allows for both automated processing and human review of damage assessment results. Moreover, the structured format supports efficient data processing, trend analysis, and maintenance prioritization while preserving the contextual richness of the original assessment.

	\section{Wind Turbine Blade Inspection Case Study}
	In this section, we examine the feasibility and effectiveness of the proposed training-free approach for wind turbine blade inspection. We will detail each step of the framework, from dataset selection and knowledge base preparation to inspection results.
	
	\subsection{Datasets overview}
\begin{figure}[htbp]
		\centering
		\includegraphics[width=\columnwidth]{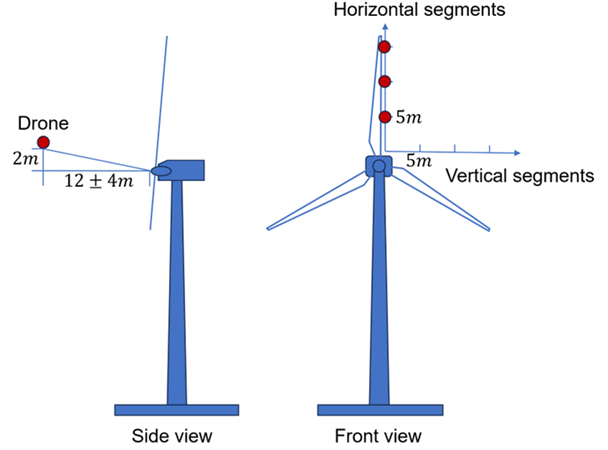}
		\caption{Schematic diagram of photography of drone for wind turbine \cite{b56}}
		\label{fig:f4}
	\end{figure}

	This study utilizes two open-source datasets: one from Chen \cite{b56,b57} and another from Foster et al \cite{b21}. The Chen dataset includes both optical and thermal imagery. All blade videos are captured using either DJI Zenmuse H20T or DJI Mavic 2 Enterprise Advanced drones while wind turbines operated normally. For thermal imaging, the fusion color palette is selected. The data collection protocol involves positioning the drone approximately 12±4 meters horizontally from the hub nose and 2 meters vertically (Figure ~\ref{fig:f4}). To avoid thermal interference from the turbine, the camera is tilted upward by 15 degrees before capturing paired optical and thermal videos. For longer blades, multiple segments horizontally or vertically are recorded, maintaining a 5-meter interval between filming positions. Videos are taken from both sides of the blades. To enhance data diversity and model robustness, footage from various angles and distances are also recorded. The final dataset comprises 36 videos, with both training and testing sets containing all video frames. Each frame has a resolution of 853×480 pixels. While thermal images were collected, they are not included in our current study. Since our goal is to achieve training-free inspection capability, we randomly selected images from both the training and testing datasets, rather than adhering to their original division. The second dataset originates from Shihavuddin and Chen \cite{b58}, with original images at 5280×2970-pixel resolution. In a recent study, Foster et al \cite{b21} subdivided these original images into 72 smaller segments of 586×371 pixels each to accommodate YOLOv5 input requirements. For our inspection, we use this more recent 586×371-pixel version. 
	
	It is important to note that neither dataset comprehensively covers all possible damage types. The first dataset primarily contains healthy state examples and crack damage. The second dataset predominantly features corrosion, erosion, surface peeling, and dirt accumulation, though Foster et al [21] simplified their classification to just two categories: dirt and damage. To broaden the range of identifiable damage types, the datasets were supplemented with three additional damaged blade images exhibiting lightning strike and burning, icing and snow accumulation, and leading-edge erosion, as shown in Figure 5 from [59]. The final evaluation dataset comprises 110 images encompassing the following conditions: healthy state, corrosion, erosion, surface peeling, icing and snow, lightning strike and burning, and cracks. These images were drawn from both datasets and cover a range of imaging conditions including daytime, nighttime, dusk, and overcast weather. As images in the second dataset often display multiple damage types simultaneously (such as combined corrosion/erosion with surface peeling), all damage types are further categorized into four mechanism-based groups: healthy state, surface damage (corrosion, erosion, surface peeling, rust), environmental damage (icing, snow, lightning strike, burning), and structural damage (fracture, cracks). Finally, there are 110 images used for testing. Evaluation metrics and the confusion matrix are reported based on these images and four categories.
		
	\begin{figure*}[htbp]
		\centering
		\includegraphics[width=\textwidth]{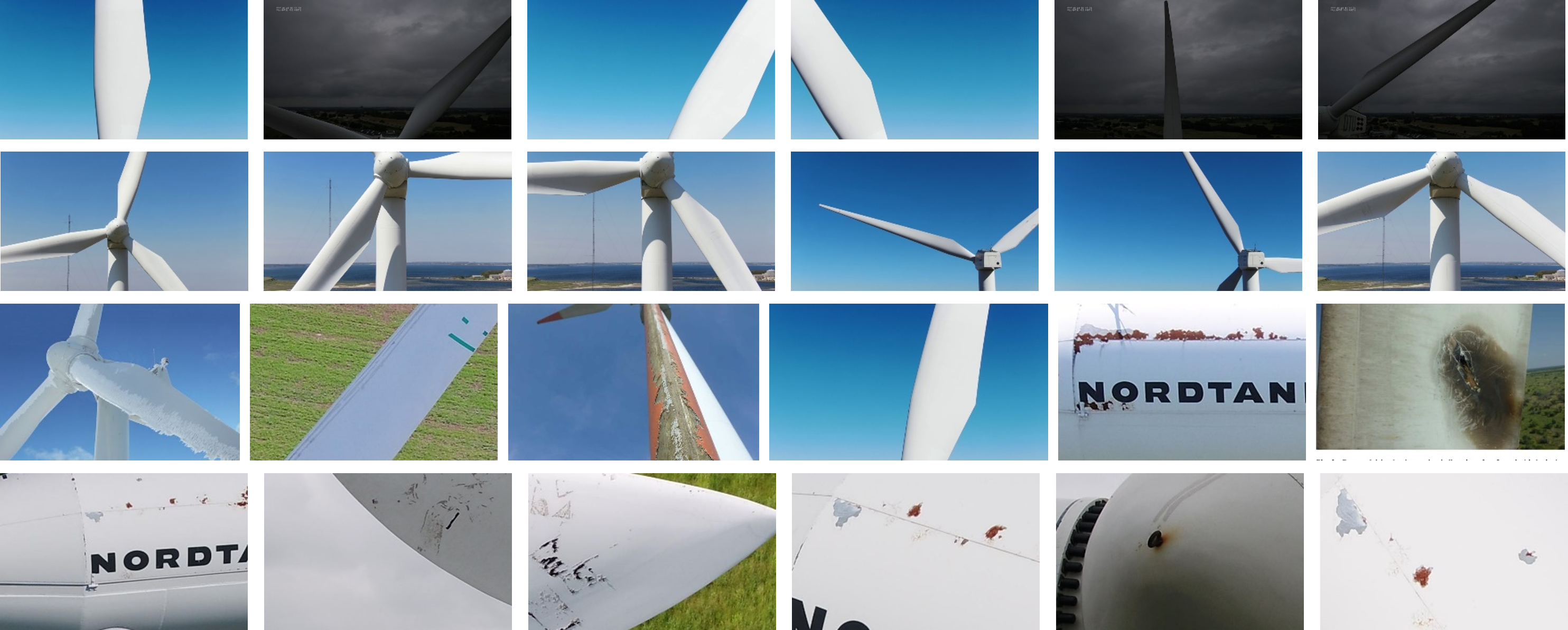}
		\caption{Wind turbine blade testing image samples from datasets \cite{b21,b56,b59}.}
		\label{fig:f5}
	\end{figure*}

	\subsection{Knowledge base preparations}
	We aim to achieve blade inspection towards training-free or fine-tuning large language models. Instead, we adopt a retrieval-augmented generation approach that relies on specialized domain knowledge. In this study, we construct four professional knowledge bases: three textual and one image–text database. The first contains descriptions of blade damage types, including visual appearance, color changes, typical locations, and other relevant cues (e.g., an example description of crack damage is given in Table ~\ref{tab:knowledge_base_examples}). The second describes turbine blade structures, covering materials, surface coatings, markings, lightning conductors, and vortex generators that should not be misclassified as damage. The third consists of simulated maintenance logs summarizing past damage cases, severity levels, and approximate classifications; although secondary to our analysis, this resource supports rough severity estimation when damage is detected. The fourth knowledge base is multimodal, pairing blade photographs under different health conditions with textual descriptions. These descriptions specify lighting and weather conditions, the number of blades visible, and the type of damage if present. Together, the four knowledge bases are embedded and stored in a vector database. During inference, similarity search with reranking retrieves the most relevant entries, which are combined with a predefined prompt. The system then uses the top three retrieved references to support interpretable blade inspection. It should be noted that the reference images included in the knowledge base are drawn only from the Chen datasets \cite{b56,b57}, meaning that part of the test images originates from different sources and are not present in the knowledge base.

	\begin{table*}[htbp]
		\centering
		\caption{Knowledge base preparations and examples.}
		\label{tab:knowledge_base_examples}
		\setlength{\tabcolsep}{3pt}
		\begin{tabular}{|p{100pt}|p{360pt}|}
			\hline
			\rowcolor{orange!30}
			\textbf{Knowledge Types} & \textbf{Sample (Part of knowledge base)} \\
			\hline
			Image/Textual & 
			\begin{minipage}{\linewidth}
				\centering
				\includegraphics[width=0.4\linewidth]{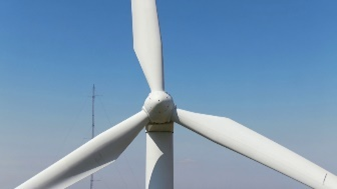}
				
				$\bullet$ The image was captured during daytime with clear blue-sky conditions. It shows three wind turbine blades with a visible crack at the end of the bottom right blade. The crack is oriented perpendicular to the longitudinal edge of the blade, which may indicate structural stress damage.
			\end{minipage} \\
			\hline
			Descriptions of damage types & 
			$\bullet$ Cracks: The obvious features for the cracks are that they are perpendicular to the length of the blade. Crack damage looks like linear fractures on the blade surface, often appearing as fine lines that can range from microscopic to several centimeters in length. They typically start at stress concentration points and may be straight, branched, or web-like. Fresh cracks appear as sharp, clean breaks with defined edges, while older cracks may have discoloration around the edges. They can be superficial (affecting only the outer layer) or structural (penetrating deeper into the blade material). \\
			\hline
			Descriptions of turbine & 
			$\bullet$ Wind turbine blades are typically made of composite materials, primarily fiberglass reinforced polyester or epoxy, and sometimes carbon fiber for larger blades.
			
			$\bullet$ A load-bearing spar or spar caps running the length of the blade.
			
			$\bullet$ Leading and trailing edge reinforcements.
			
			$\bullet$ Outer shell or skin made of composite materials.
			
			$\bullet$ Protective coating and paint to shield against environmental elements.
			
			$\bullet$ Root section reinforced with metal for connection to the hub.
			
			$\bullet$ There is usually a seam line on the back side of the blade, along the length direction but it is not damage. \\
			\hline
			Maintenance logs & 
			$\bullet$ Level 1 (Minor): Superficial damage that does not affect structural integrity or performance. Examples include minor surface erosion, small scratches, or minor coating damage. Monitoring recommended.
			
			$\bullet$ Level 2 (Low): Early-stage damage that may progress if not addressed but does not present immediate concern. Examples include small cracks less than 10cm, early stage leading edge erosion, or limited surface peeling. Scheduled repair recommended within 3-6 months. \\
			\hline
		\end{tabular}
	\end{table*}

	\subsection{Identification Results}
\begin{figure*}[!t]
		\centering
		\subfloat[]{\includegraphics[width=2.5in]{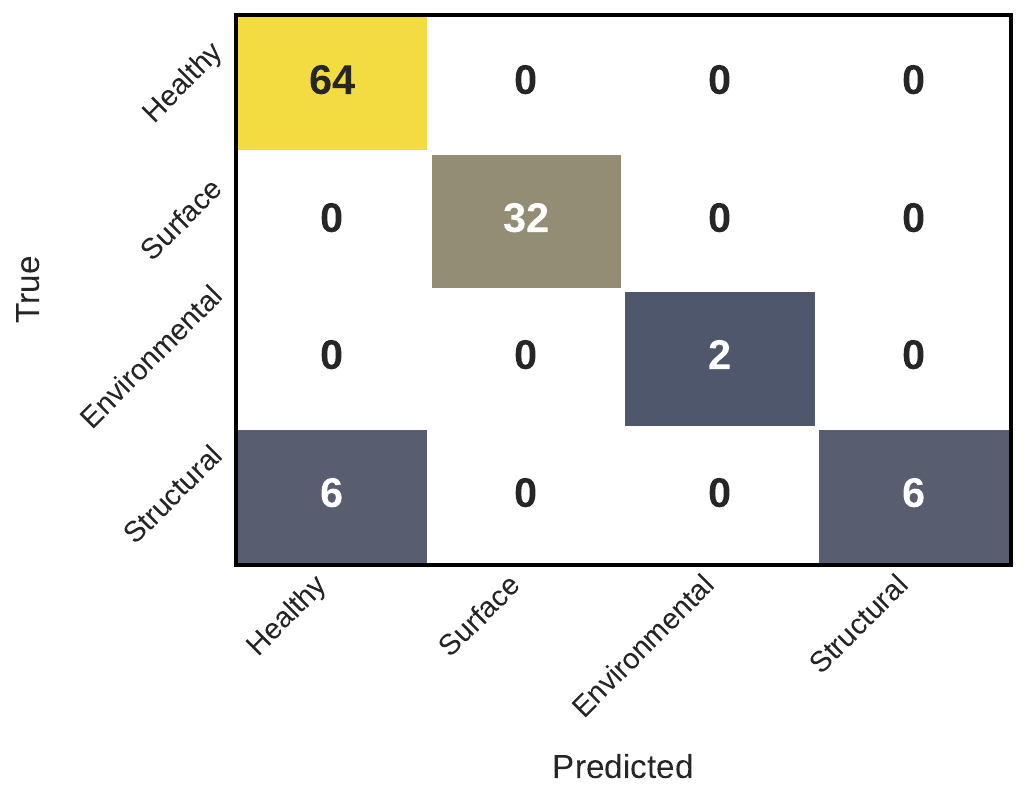}%
			\label{fig:f6a}}
		\hfil
		\subfloat[]{\includegraphics[width=2.5in]{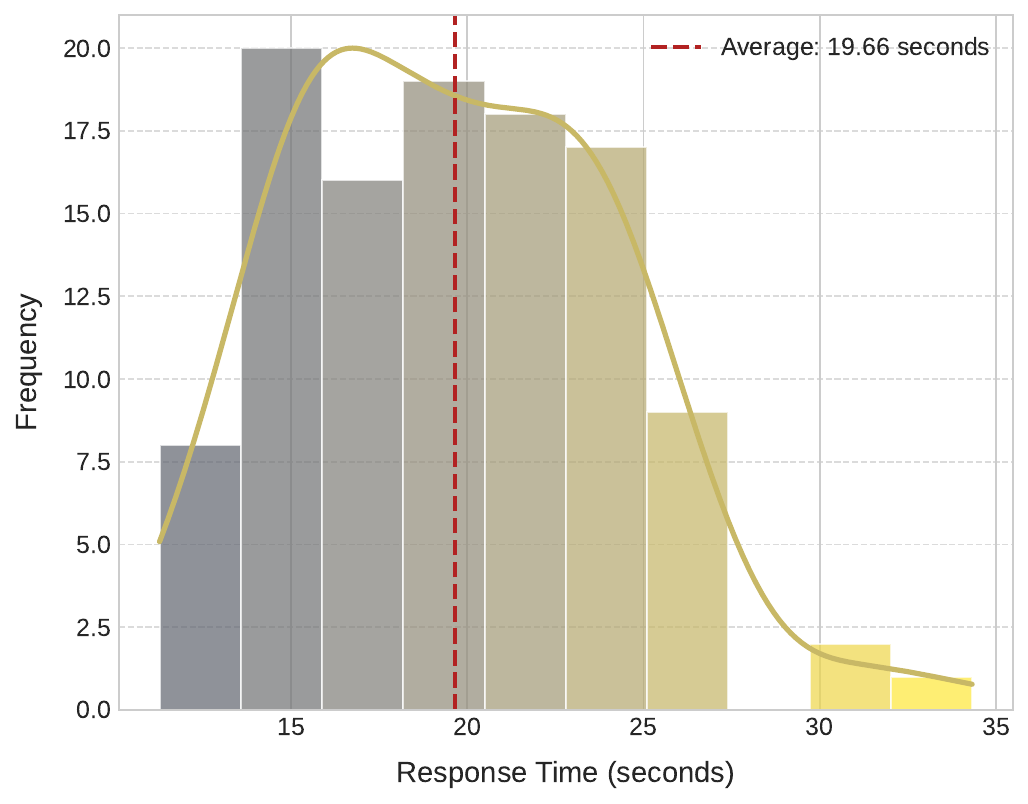}%
			\label{fig:f6b}}
		\caption{Result comparisons: (a) confusion matrix and (b) response time distribution.}
		\label{fig:f6}
	\end{figure*}

	It is worth mentioning that our approach follows a setting towards training-free: the RAG-grounded VLM is not trained or fine-tuned on the target dataset, but directly applied to unseen images. Therefore, concepts such as cross-validation or train/test splits are not applicable in the conventional sense. Our evaluation is performed on 110 independent test images from open-source datasets, with ground-truth labels established by domain experts. After processing through the RAG-grounded VLM, we extract the model judgments from the originally generated responses and compare them with the actual damage conditions. The results are shown in Figure~\ref{fig:f6}. From the confusion matrix in Figure~\ref{fig:f6a}, we can see that the proposed approach can accurately identify different types of damage. The method achieves an overall accuracy of 94.55\%, with precision of 0.9786, recall of 0.8750, and F1 score of 0.9055. The framework correctly classifies all Healthy and Surface damage instances, and both Environmental damage cases are also identified without error. The primary source of misclassification lies in Structural damage, where 6 out of 12 crack images are predicted as Healthy. Most of these misclassified cases are captured under poor lighting conditions, where crack features are visually obscured and difficult to distinguish from an intact blade surface even for human inspectors.
	
	Another noteworthy aspect is the distribution of response times, as shown in Figure~\ref{fig:f6b}. Across the 110 test images, the average response time was 19.66 seconds on Google Colab equipped with 12.7GB system RAM and a T4 GPU with 15.0GB memory. This end-to-end latency includes several components: approximately 2--4 seconds for cloud storage access, 8--12 seconds for knowledge base retrieval, 6--10 seconds for VLM inference, and 3--5 seconds for structured response generation. Most responses cluster around 20 seconds, while a few are completed in as little as 10 seconds. Cloud I/O and retrieval currently account for nearly half of the total time, suggesting that local knowledge base deployment or improved caching strategies could substantially reduce latency. Similarly, inference time can be shortened by adopting model distillation or quantization techniques, or by deploying dedicated GPU resources. Preliminary tests suggest that such optimizations could reduce the response time to 8--12 seconds per image.
	
	Table ~\ref{tab:raw_responses_example} presents two complete examples of generation process of query results for several different damage images. The RAG-grounded VLM provides formatted responses according to the user and prompt. When answering whether damage is present, clear judgments are required. Additionally, based on the description in the fourth question, we can see from the text how the model observed the damage, what it looks like, and thereby made reasonable judgments based on the observations.
	
	\begin{table*}[t]
		\centering
		\caption{Examples of raw responses by VLM on test data.}
		\label{tab:raw_responses_example}
		\setlength{\tabcolsep}{3pt}
		\begin{tabular}{|p{460pt}|}
			\hline
			\rowcolor{orange!30}
			\textbf{Examples for Test Image Data Responses} \\
			\hline
			\textbf{Image Index}: 1
			
			\textbf{Ground Truth}: healthy
			
			\textbf{Raw Response: ``Based on the provided image and the reference information, here is the analysis:}
			
			\textbf{1. How many blades are visible in the image?}
			
			- One blade is visible in the image.
			
			\textbf{2. Is there visible damage on any of the turbine blades in the image?}
			
			- No visible damage is observed on the turbine blade in the image.
			
			\textbf{3. If yes, what specific type of damage can be identified in this damage lists...}
			
			- No specific type of damage is identified.
			
			\textbf{4. Provide a detailed description of the damage observed...}
			
			- The blade appears to be in good condition with no visible signs of damage such as cracks, corrosion, erosion, rust, delamination, fractures, dents, ice, snow, surface peeling, wear, or lightning strikes/burning. The clear line along the longitudinal direction is a normal manufacturing feature rather than damage.
			
			\textbf{5. Rate the severity of the damage on a scale of 1-5, where 1 is minor and 5 is severe.}
			
			- Severity rating: 1 (minor) since no damage is observed.''
			
			\textbf{Extracted Type}: Healthy
			
			\textbf{Extracted Severity}: 0
			
			\textbf{Context Used}: Text Docs: img\_h10\_chunk\_0, img\_h5\_chunk\_0, img\_h15\_chunk\_0; Image Docs: img\_h14, img\_h13, img\_crk2-3 \\
			\hline
			\textbf{Image Index}: 6
			
			\textbf{Ground Truth}: ice/snow
			
			\textbf{Raw Response: ``Based on the provided image and the reference information, here is the analysis:}
			
			\textbf{1. Number of Blades Visible}: The image shows three blades of a wind turbine.
			
			\textbf{2. Visible Damage}: There is visible damage on one of the turbine blades.
			
			\textbf{3. Specific Type of Damage}: The damage observed appears to be ice accumulation along the edge of one of the blades. This does not match the specific types of damage listed but it is closest to ``Ice'' and ``Snow''.
			
			\textbf{4. Detailed Description of Damage}: The blade in question has a significant accumulation of ice along its edge. The ice formation is uneven and extends along the length of the blade, which can affect the aerodynamic performance of the turbine. The ice appears thick and jagged, indicating that it has built up over time due to cold and wet conditions.
			
			\textbf{5. Severity Rating}: On a scale of 1-5, where 1 is minor and 5 is severe, the severity of the ice accumulation would be rated as a 4.''
			
			\textbf{Extracted Type}: Wear, Ice
			
			\textbf{Extracted Severity}: 4
			
			\textbf{Context Used}: Text Docs: img\_Unspecified\_h10\_chunk\_0, img\_Unspecified\_h5\_chunk\_0, img\_Unspecified\_h15\_chunk\_0; Image Docs: img\_Crack\_crk1-3, img\_Crack\_crk1-7, img\_Unspecified\_hole1 \\
			\hline
		\end{tabular}
	\end{table*}

	\subsection{Knowledge usage analysis}
\begin{figure*}[htbp]
		\centering
		\includegraphics[width=\textwidth]{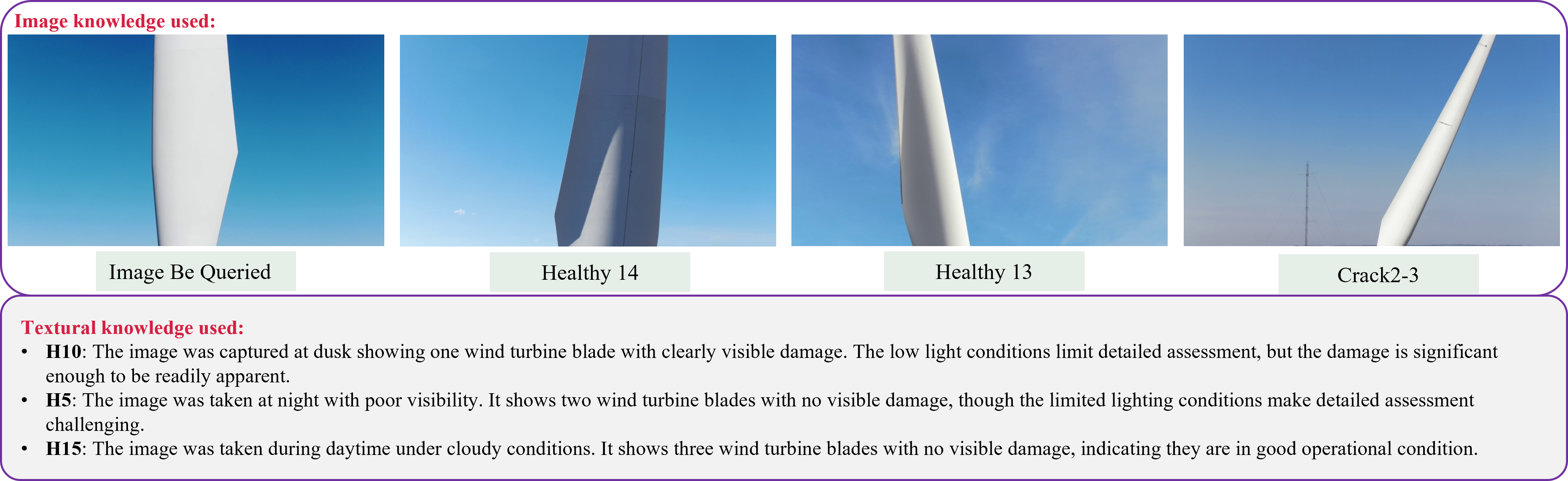}
		\caption{Knowledge usages of testing image 1.}
		\label{fig:f8}
	\end{figure*}

	For each image analysis, we track which knowledge base components are used as references. As illustrated in Table ~\ref{tab:raw_responses_example} and Figure ~\ref{fig:f8}, the system selects the three most relevant text chunks and three reference images to support its assessment. In one example (Marked as index 1 in Table ~\ref{tab:raw_responses_example}), the retrieved texts describe blades under different lighting conditions (e.g., dusk with visible damage, night with poor visibility, and cloudy daytime with no damage), helping the model account for contextual challenges. The visual references include two healthy blades and one cracked blade, which provide concrete examples for comparison. By combining these textual and visual cues, the RAG system grounds its reasoning in complementary knowledge sources and determines whether the query image more closely resembles a healthy blade or exhibits features of a specific damage type.
	
	This hybrid approach to knowledge retrieval offers several advantages. The textual knowledge provides valuable context about damage types, severity levels, and assessment challenges. The image references enable direct visual comparison, helping the model identify similar patterns or anomalies. By incorporating references from various lighting conditions, the system can better handle images taken in suboptimal environments. By tracking which knowledge sources were used, we gain insight into how the model arrived at its conclusions, making the system more transparent and trustworthy. Analysis of knowledge usage across our test dataset revealed that certain reference images and text descriptions are consistently retrieved for specific damage types, suggesting that the RAG system effectively identified visual and contextual patterns relevant to accurate damage assessment. This consistency in knowledge retrieval also indicates that our embedding approach successfully captures both semantic meaning in text and visual features in images that are relevant to the turbine blade damage detection task.
	
	\section{Ablation Study and Interpretation}
	\subsection{Retrieval Component Ablation}
	
\begin{figure*}[!t]
		\centering
		\subfloat[]{\includegraphics[width=2.8in]{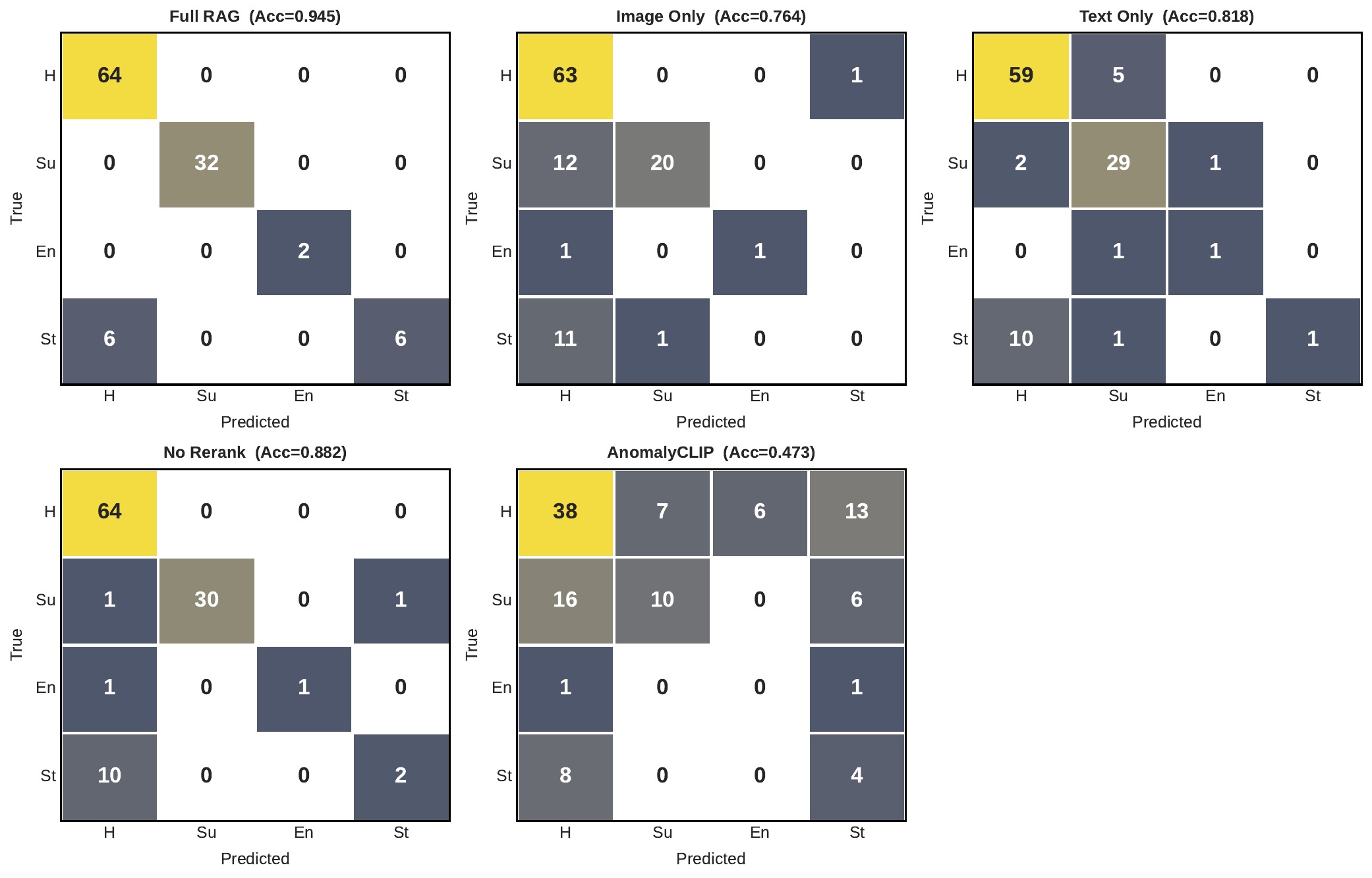}%
			\label{fig:cm_all}}
		\hfil
		\subfloat[]{\includegraphics[width=2.8in]{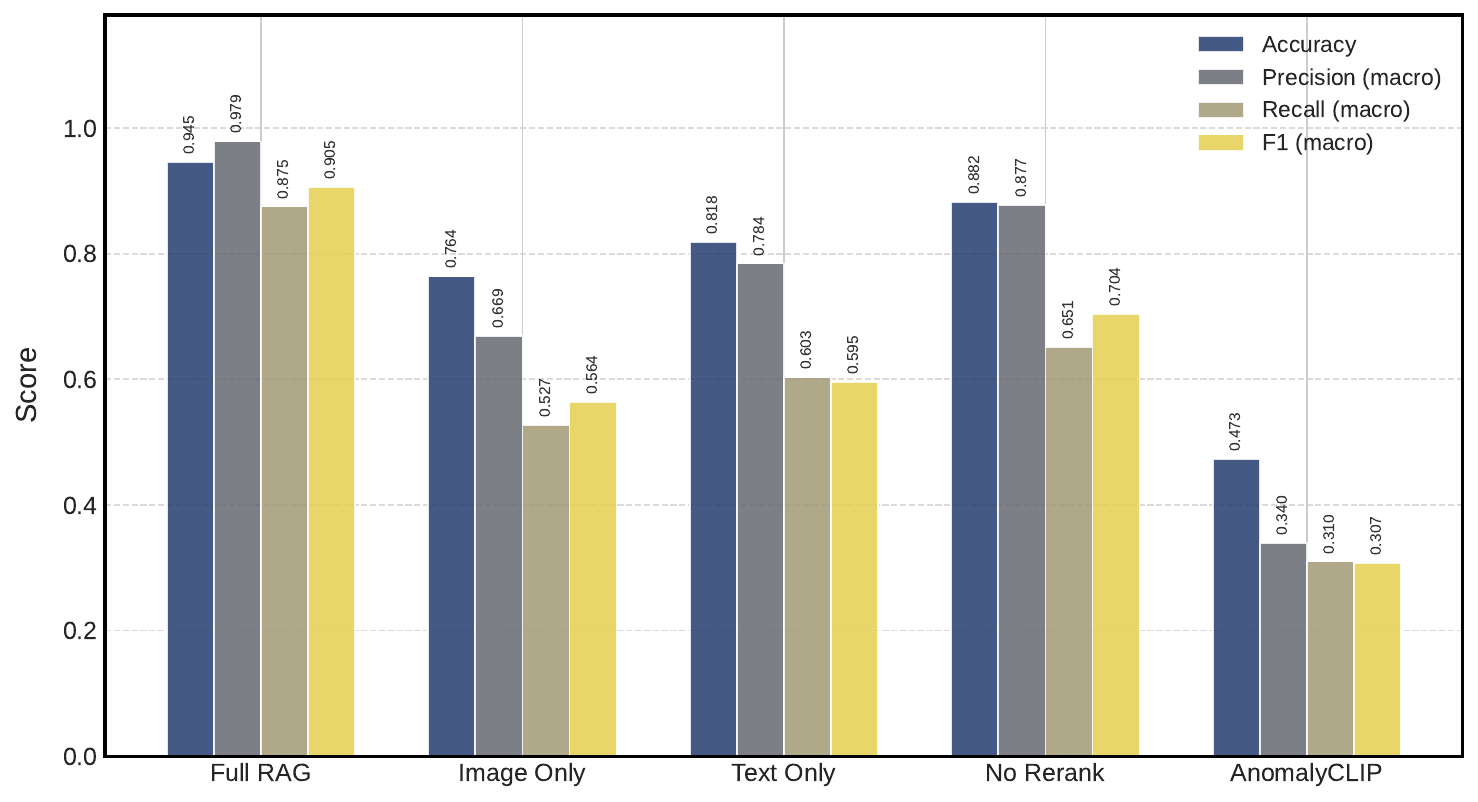}%
			\label{fig:metrics_bar}}
		\caption{Result comparisons: (a)confusion matrices and (b) metric comparisons.}
		\label{fig:f_abolation}
	\end{figure*}

	To systematically assess the contribution of each retrieval component, we evaluate three ablated configurations against the full system: (1) Image Only, which retrieves only visual reference images without incorporating textual documents; (2) Text Only, which relies solely on textual document retrieval without image reference matching; and (3) No Rerank, which applies the full hybrid similarity search across both modalities but omits the keyword-aware reranking step. All configurations use the same Qwen-VL-Max backbone and the same prompt structure, and are evaluated on the identical 110-image test set. The confusion matrices and overall metrics are shown in Figure~\ref{fig:cm_all} and Figure~\ref{fig:metrics_bar}, respectively.
	
	The results show that the Full RAG system achieves the highest overall performance, with an accuracy of 94.55\% and a macro-averaged F1 of 0.9055. When image retrieval is removed, the Text Only shows degraded performance on Environmental damage (F1 = 0.500). This is not surprising: damage types such as icing and lightning strike/burning carry strong visual signatures,which are distinctive color patterns and surface textures, that are difficult to convey through text descriptions alone. Without concrete reference images to ground the VLM's visual observations, the model cannot reliably associate what it sees with the correct damage category. Conversely, removing textual retrieval in the Image Only variant leads to complete failure on Structural damage (F1 = 0.000). Crack damage is inherently visually heterogeneous, as its appearance varies significantly with blade material, crack stage, and imaging conditions, making image-level similarity search an unreliable guide. The textual knowledge base compensates for this by providing explicit descriptors such as crack orientation, edge sharpness, and typical spatial locations, which anchor the VLM's reasoning when visual cues alone are ambiguous.
	
	Removing the reranking step has a more subtle but consistent effect. The No Rerank variant maintains competitive accuracy at 88.18\%, but the retrieved context is ordered purely by embedding similarity, which does not always surface the most task-relevant documents. The keyword-aware reranking explicitly promotes documents containing query-relevant terms, ensuring that the VLM receives more targeted evidence for each inspection case. The cumulative effect of this refinement is reflected in the overall F1 gap between No Rerank (0.7038) and Full RAG (0.9055).
	
	\subsection{Comparison with Zero-Shot Vision-Language Baseline}
	To contextualize the proposed framework against existing zero-shot vision-language approaches, we compare against AnomalyCLIP [31], a representative method that employs object-agnostic prompt learning with a CLIP backbone to detect anomalies across diverse domains without task-specific training. AnomalyCLIP is evaluated on the same 110-image test set under the same four-class taxonomy. Predicted categories are derived from the AnomalyCLIP classification output, and results are reported using the same per-class precision, recall, and F1 metrics.
	
	As shown in Table~\ref{tab:perclass} and Figure~\ref{fig:metrics_bar} , AnomalyCLIP achieves an overall accuracy of 47.27\% and a macro-averaged F1 of 0.307, substantially below all RAG-grounded variants. The method produces reasonable results on the Healthy category (F1 = 0.598) but performs poorly on Surface damage (F1 = 0.408) and fails entirely on Environmental damage (F1 = 0.000). The confusion matrix in Figure~\ref{fig:cm_all} reveals a strong systematic bias toward over-predicting Structural damage: AnomalyCLIP assigns 13 Healthy images and 6 Surface damage images to the Structural category, resulting in a high false positive rate for this class. This behavior reflects a fundamental limitation of general-purpose anomaly detection models when applied to domain-specific industrial inspection tasks: without structured access to domain knowledge describing what constitutes each damage type, the model cannot distinguish between category-level visual signatures that require expert context to interpret.
	
	\begin{table}[!t]
		\centering
		\caption{Per-class classification metrics for all evaluated methods (n = 110).}
		\label{tab:perclass}
		\begin{tabular}{llccc}
			\hline
			\textbf{Method} & \textbf{Class} & \textbf{Precision} & \textbf{Recall} & \textbf{F1} \\
			\hline
			\multirow{4}{*}{Full RAG}
			& Healthy       & 0.9143 & 1.0000 & 0.9552 \\
			& Surface       & 1.0000 & 1.0000 & 1.0000 \\
			& Environmental & 1.0000 & 1.0000 & 1.0000 \\
			& Structural    & 1.0000 & 0.5000 & 0.6667 \\
			\hline
			\multirow{4}{*}{Image Only}
			& Healthy       & 0.7241 & 0.9844 & 0.8344 \\
			& Surface       & 0.9524 & 0.6250 & 0.7547 \\
			& Environmental & 1.0000 & 0.5000 & 0.6667 \\
			& Structural    & 0.0000 & 0.0000 & 0.0000 \\
			\hline
			\multirow{4}{*}{Text Only}
			& Healthy       & 0.8310 & 0.9219 & 0.8741 \\
			& Surface       & 0.8056 & 0.9062 & 0.8529 \\
			& Environmental & 0.5000 & 0.5000 & 0.5000 \\
			& Structural    & 1.0000 & 0.0833 & 0.1538 \\
			\hline
			\multirow{4}{*}{No Rerank}
			& Healthy       & 0.8421 & 1.0000 & 0.9143 \\
			& Surface       & 1.0000 & 0.9375 & 0.9677 \\
			& Environmental & 1.0000 & 0.5000 & 0.6667 \\
			& Structural    & 0.6667 & 0.1667 & 0.2667 \\
			\hline
			\multirow{4}{*}{AnomalyCLIP}
			& Healthy       & 0.6032 & 0.5938 & 0.5984 \\
			& Surface       & 0.5882 & 0.3125 & 0.4082 \\
			& Environmental & 0.0000 & 0.0000 & 0.0000 \\
			& Structural    & 0.1667 & 0.3333 & 0.2222 \\
			\hline
		\end{tabular}
	\end{table}

	In contrast, the proposed RAG-grounded VLM framework injects domain-specific knowledge at inference time, enabling the model to ground its visual analysis in concrete damage descriptions and reference imagery. This knowledge grounding is particularly consequential for Environmental and Surface damage, where the visual appearance of the damage is strongly context-dependent and may overlap with normal blade surface features. These results confirm that structured domain knowledge integration, rather than visual feature extraction capability alone, is the key differentiating factor for accurate multi-class blade damage classification.

	\begin{figure*}[!t]
		\centering
		\includegraphics[width=0.7\textwidth]{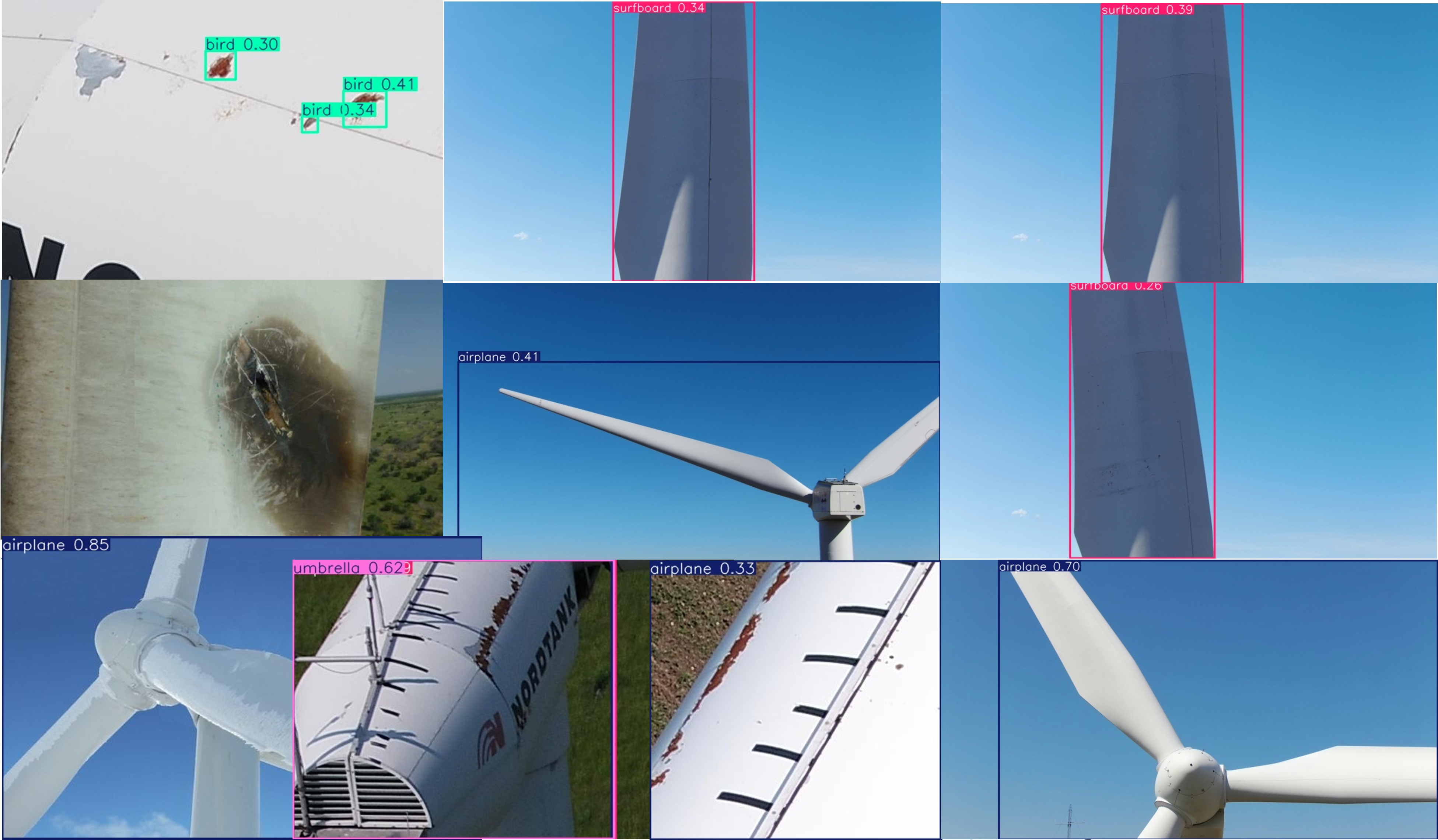}
		\caption{Representative detection results from YOLO-World (zero-shot).}
		\label{fig:villina}
	\end{figure*}
	
	\begin{figure*}[!t]
		\centering
		\includegraphics[width=0.7\textwidth]{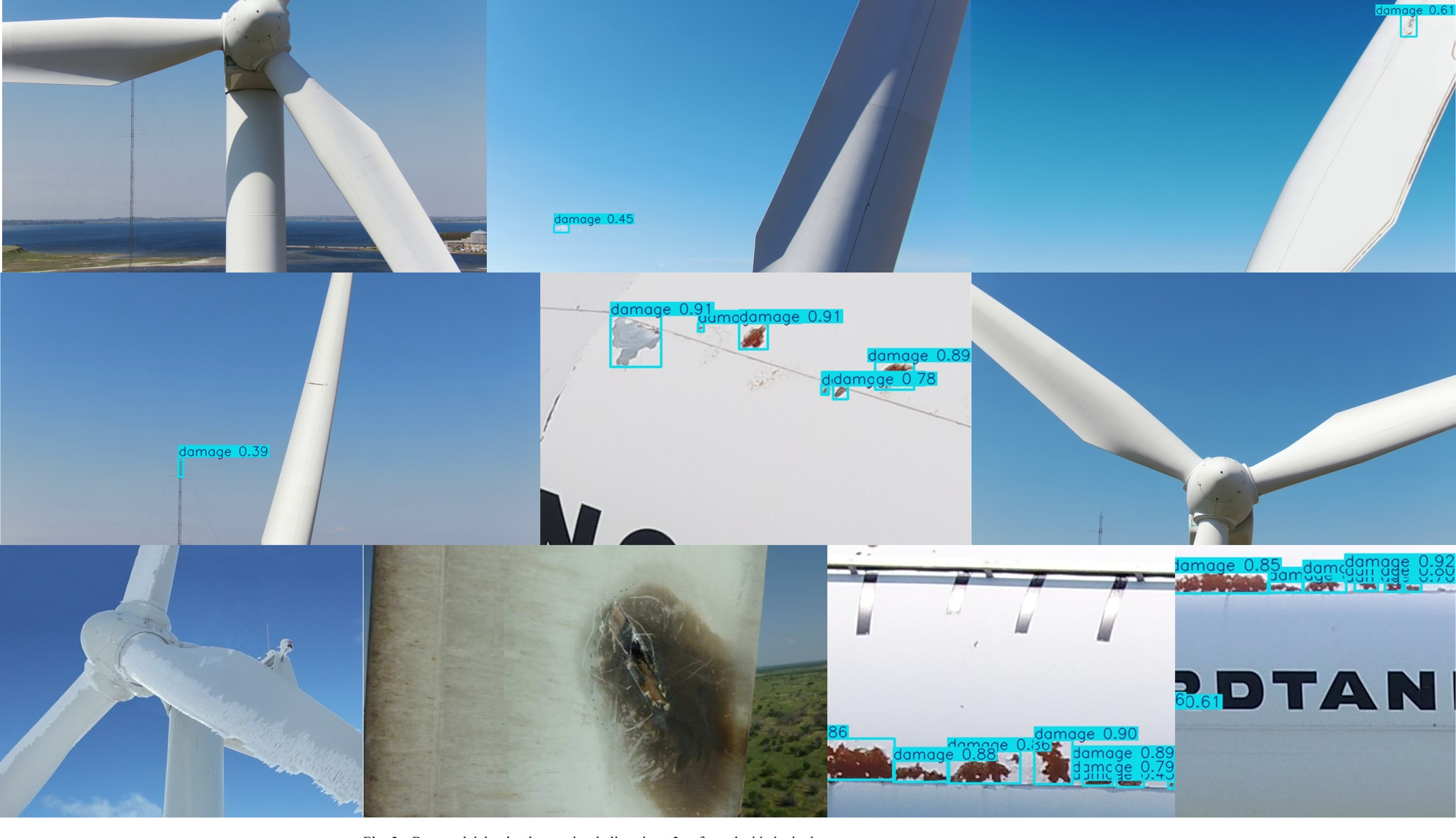}
		\caption{Representative detection results from the retrained YOLOv8n. Detections are largely confined to surface damage patterns from the training distribution, with structural and environmental damage categories largely missed.}
		\label{fig:retraining}
	\end{figure*}
	
	\begin{figure*}[!t]
		\centering
		\includegraphics[width=0.6\textwidth]{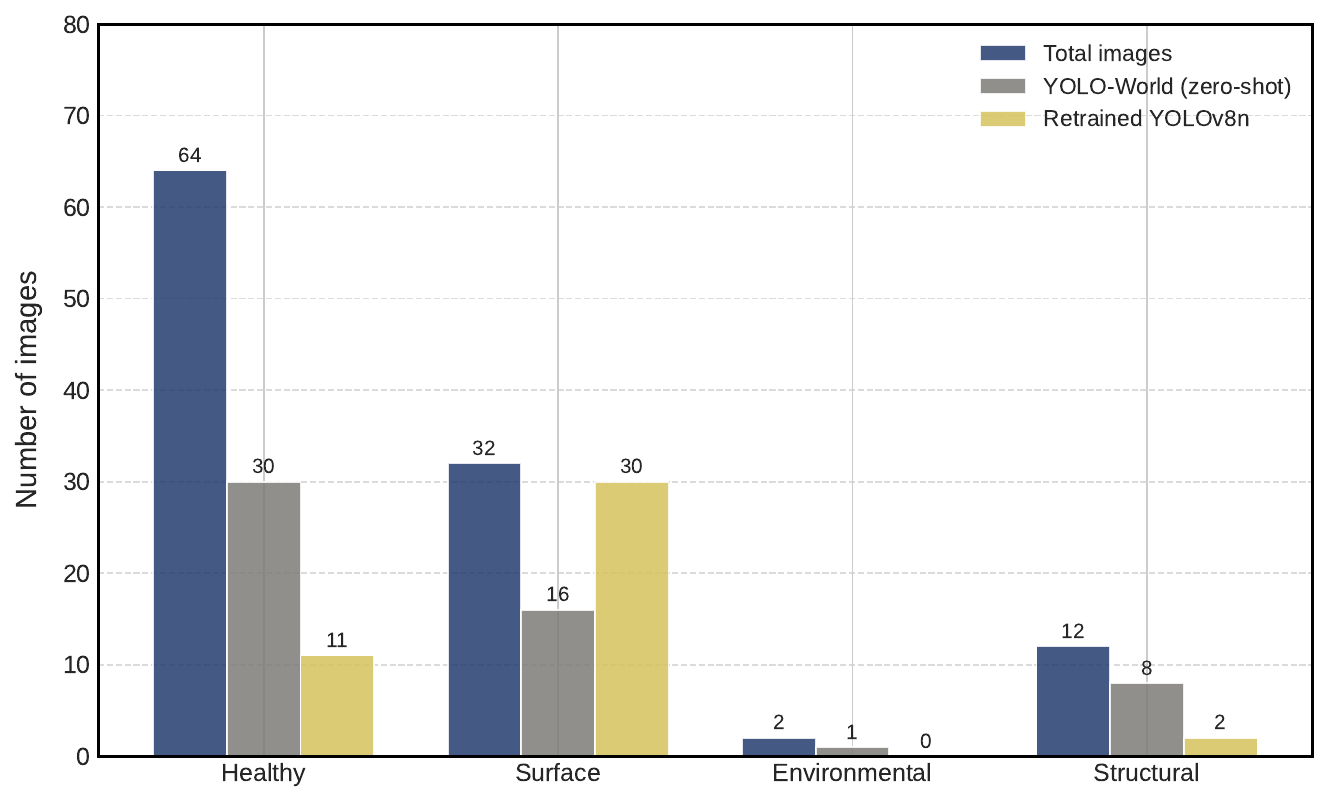}
		\caption{Detection count by ground-truth damage category for YOLO-World and retrained YOLOv8n across all 110 test images.}
		\label{fig:f14}
	\end{figure*}
	
	\subsection{Performance Comparisons to YOLO Models}
	
	We next compare our RAG-grounded VLM with YOLO-based detectors to highlight the advantages of retrieval-augmented, training-free inspection over conventional supervised learning methods. YOLO~[60] is chosen as the representative baseline because it is one of the most widely adopted object detection frameworks, and its zero-shot extension (YOLO-World) is directly aligned with our inspection scenario. Unlike anomaly detection models that primarily target texture-level surface irregularities, YOLO is designed for object- and region-level detection, making it more suitable for large structural components such as wind turbine blades where damage often occurs at multiple scales.
	We first evaluate YOLOv8s-WorldV2 in a zero-shot setting, applying it directly to the 110 test images without any task-specific training. Representative detection results are shown in Figure~\ref{fig:villina}. The model fails at both the object recognition and damage detection levels: turbine blades are misidentified as airplanes and surfboards, surface corrosion spots are labeled as birds, and blade structures are assigned to categories such as umbrellas. Across the full test set, 55 out of 110 images receive at least one detection output, yet 30 of these occur on ground-truth Healthy images, indicating a high false positive rate. None of the detections correspond to meaningful damage identification. This outcome reflects a fundamental semantic mismatch: YOLO-World is pretrained on broad community datasets with generic object categories, and industrial damage detection requires domain-specific semantic granularity that general-purpose pretraining cannot provide.
	
	We then retrain a YOLOv8n model using 3000 images from dataset 2 under full supervision, with two damage categories: damage and dirt. Representative results are shown in Figure~\ref{fig:retraining}. The retrained model successfully detects surface-level damage in images similar to its training distribution, producing bounding boxes with high confidence scores on corrosion and peeling patterns. However, its generalization is severely limited. Out of 32 Surface damage images, 30 receive valid detections. In sharp contrast, only 2 out of 12 Structural damage images are detected, and none of the 2 Environmental damage cases are identified. Moreover, 11 Healthy images receive false positive detections, where normal blade surface features such as manufacturing seams and structural markings are incorrectly flagged as damage. The quantitative breakdown across all damage categories is summarized in Figure~\ref{fig:f14}.
	
	The quantitative comparison reveals a consistent pattern. The retrained model performs well on the category it was trained on, but collapses on categories outside its training distribution. Structural damage, which is visually distinct from surface corrosion and requires understanding of crack geometry and orientation, is almost entirely missed. Environmental damage, which involves scene-level cues such as ice accumulation along blade edges, is not part of the training taxonomy and is therefore invisible to the model. Standard detection metrics such as mAP are not reported, as the near-zero valid detection rate on cross-domain images precludes meaningful quantitative comparison. The failure pattern itself is the finding.
	
	This is precisely the limitation that the proposed framework is designed to address. Supervised pipelines are caught in a continuous cycle of data collection, annotation, and retraining whenever new damage types emerge or imaging conditions change. A crack dataset produces a crack detector; a corrosion dataset produces a corrosion detector — and neither transfers to the other without retraining. The RAG-grounded VLM breaks this cycle. Domain knowledge for new damage types is incorporated by updating the knowledge base, with no model adaptation required. The framework generalizes across datasets, damage categories, and imaging conditions by grounding its visual reasoning in retrieved expert knowledge rather than memorized training patterns. Although the embedding models inherently leverage broad pretraining corpora, the inspection task itself remains training-free under the accepted definition, where this term denotes the absence of task-specific model adaptation. The modular design further supports transfer to related industrial domains such as bridge inspection, corrosion monitoring, and manufacturing defect detection, highlighting the broader generalizability of the approach.

	\section{Conclusion}
	This study introduces a training-free, knowledge-grounded framework for wind turbine blade inspection by integrating Retrieval-Augmented Generation with Vision Language Models. Evaluated on 110 blade images across four damage categories, the Full RAG framework achieves an overall accuracy of 94.55\%, with macro-averaged precision of 0.9786, recall of 0.8750, and F1 score of 0.9055. Ablation studies confirm that both the hybrid text--image retrieval mechanism and the keyword-aware reranking step are individually necessary: removing image retrieval causes complete failure on structural damage, while removing textual retrieval leads to systematic misclassification of environmental damage. Comparison against AnomalyCLIP and YOLO-based detectors further validates that structured domain knowledge integration is the key differentiating factor, and that supervised pipelines remain fundamentally constrained by their training distribution in cross-domain inspection scenarios. The primary limitation of the current framework is the detection of structural damage under poor lighting conditions, where crack features remain visually subtle regardless of retrieval strategy. Future work will focus on knowledge base augmentation with diverse crack imagery, integration of bounding box localization for spatial damage assessment, and uncertainty-guided knowledge updating to improve robustness in real-world deployment. Through the combination of retrieval-augmented generation and vision language models, this work contributes a data-efficient and interpretable solution for industrial inspection that reduces dependence on labeled datasets while ensuring critical infrastructure safety.
	
	\section*{Acknowledgment}
	Portions of this article were refined using ChatGPT (OpenAI), Calude (Anthropic) for language improvement and clarity.

	\newpage
	
	\begin{IEEEbiography}[{\includegraphics[width=1in,height=1.25in,clip,keepaspectratio]{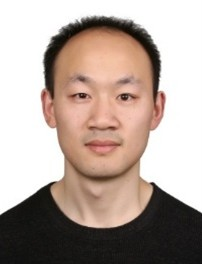}}]{Yang Zhang} (Member, IEEE) is a postdoctoral researcher at the University of Connecticut. He received his Ph.D. in Mechanical, Aerospace, and Manufacturing Engineering from the University of Connecticut, Storrs, CT, USA. His research interests include prognostics and health management, inverse analysis and multi-objective optimization, mechatronic system synthesis, and AI-driven engineering applications, with a recent focus on intelligent systems, automation, and advanced manufacturing enabled by language models and robotics.
	\end{IEEEbiography}
	
	\begin{IEEEbiography}[{\includegraphics[width=1in,height=1.25in,clip,keepaspectratio]{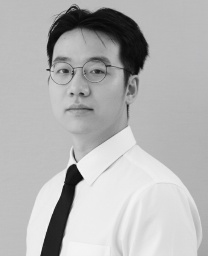}}]{Qianyu Zhou} received his PhD degree in Mechanical Engineering from the University of Connecticut. He is currently engaged in research on intelligent inspection and monitoring for advanced manufacturing systems, with a focus on welding process analysis and quality prediction. His research interests include data driven modeling, multimodal sensing, and physics informed machine learning, particularly for automated surface inspection and robust learning under limited and imperfectly labeled data.
	\end{IEEEbiography}
	
	\begin{IEEEbiography}[{\includegraphics[width=1in,height=1.25in,clip,keepaspectratio]{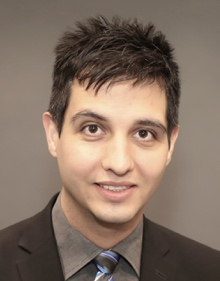}}]{Farhad Imani} is an Assistant Professor in the School of Mechanical, Aerospace, and Manufacturing Engineering at the University of Connecticut, Storrs, CT, USA. He received the B.S. degree in Industrial Engineering from the University of Science and Culture, Tehran, Iran, in 2010, the M.S. degree in Industrial Engineering from the University of Louisville, Louisville, KY, USA, in 2016, and the Ph.D. degree in Industrial Engineering and Operations Research from The Pennsylvania State University, State College, PA, USA, in 2020. His research interests include robotics and physical AI, edge computing, and smart manufacturing. He has served as an Associate Editor for the ASME Journal of Autonomous Vehicles and Systems and Elsevier Results in Control and Optimization.
	\end{IEEEbiography}
	
	\begin{IEEEbiography}[{\includegraphics[width=1in,height=1.25in,clip,keepaspectratio]{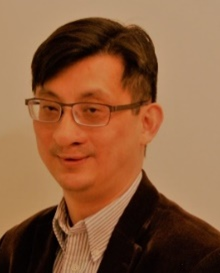}}]{Jiong Tang} (M' 09) is the Pratt \& Whitney Endowed Chair Professor in School of Mechanical, Aerospace, and Manufacturing Engineering, University of Connecticut. He received the B.S. and M.S. degrees in Applied Mechanics from Fudan University, China, in 1989 and 1992, respectively, and the Ph.D. degree in Mechanical Engineering from the Pennsylvania State University in 2001. His research interests encompass dynamics, sensing, control, automation and ML/AI. He has served as a Technical Editor for ASME/IEEE Transactions on Mechatronics, and as an Associate Editor for IEEE Transactions on Instrumentation and Measurement.
	\end{IEEEbiography}

	\EOD
	

\begin{thebibliography}{00}
		
		\bibitem{b1} L. Mishnaevsky Jr, C. B. Hasager, C. Bak, A. M. Tilg, J. I. Bech, S. D. Rad and S. Fæster, ``Leading edge erosion of wind turbine blades: Understanding, prevention and protection,'' \emph{Renewable Energy}, vol. 169, pp. 953--969, 2021.
		
		\bibitem{b2} K. Kong, K. Dyer, C. Payne, I. Hamerton and P. M. Weaver, ``Progress and trends in damage detection methods, maintenance, and data-driven monitoring of wind turbine blades–A review,'' \emph{Renewable Energy Focus}, vol. 44, pp. 390--412, 2023.
		
		\bibitem{b3} I. Gohar, W. K. Yew, A. Halimi and J. See, ``Review of state-of-the-art surface defect detection on wind turbine blades through aerial imagery: Challenges and recommendations,'' \emph{Engineering Applications of Artificial Intelligence}, vol. 144, p. 109970, 2025.
		
		\bibitem{b4} L. Jiang, S. P. Zhang, G. Q. Shen and L. Zhou, ``Acoustic Emission-based wind turbine blade icing monitoring using deep learning technology,'' \emph{Renewable Energy}, p. 122980, 2025.
		
		\bibitem{b5} M. Meng, Y. J. Chua, E. Wouterson and C. P. K. Ong, ``Ultrasonic signal classification and imaging system for composite materials via deep convolutional neural networks,'' \emph{Neurocomputing}, vol. 257, pp. 128--135, 2017.
		
		\bibitem{b6} J. Mendikute, I. Carmona, I. Aizpurua, I. Bediaga, I. Castro, L. Galdos and J. L. Lanzagorta, ``Defect detection in wind turbine blades applying convolutional neural networks to ultrasonic testing,'' \emph{NDT \& E International}, vol. 154, p. 103359, 2025.
		
		\bibitem{b7} C. Wang and Y. Gu, ``Research on infrared nondestructive detection of small wind turbine blades,'' \emph{Results in Engineering}, vol. 15, p. 100570, 2022.
		
		\bibitem{b8} M. Memari, P. Shakya, M. Shekaramiz, A. C. Seibi and M. A. S. Masoum, ``Review on the advancements in wind turbine blade inspection: Integrating drone and deep learning technologies for enhanced defect detection,'' \emph{IEEE Access}, vol. 12, pp. 33236--33282, 2024.
		
		\bibitem{b9} K. Masita, A. N. Hasan, T. Shongwe and H. A. Hilal, ``Deep Learning in Defect Detection of Wind Turbine Blades: A Review,'' \emph{IEEE Access}, vol. 13, pp. 98399--98425, 2025.
		
		\bibitem{b10} C. Yang, H. Zhou, X. Liu, Y. Ke, B. Gao, M. Grzegorzek, Z.Boukhers, T. Chen and J. See, ``BladeView: Toward automatic wind turbine inspection with unmanned aerial vehicle,'' \emph{IEEE Transactions on Automation Science and Engineering}, 22, pp.7530-7545, 2024.
		
		\bibitem{b11} L. Liu, P. Li, D. Wang and S. Zhu, ``A wind turbine damage detection algorithm designed based on YOLOv8,'' \emph{Applied Soft Computing}, vol. 154, p. 111364, 2024.
		
		\bibitem{b12} Z. Wu, Y. Zhang, X. Wang, H. Li, Y. Sun and G. Wang, ``Algorithm for detecting surface defects in wind turbines based on a lightweight YOLO model,'' \emph{Scientific Reports}, vol. 14, no. 1, p. 24558, 2024.
		
		\bibitem{b13} X. Hang, X. Zhu, X. Gao, Y. Wang and L. Liu, ``Study on crack monitoring method of wind turbine blade based on AI model: Integration of classification, detection, segmentation and fault level evaluation,'' \emph{Renewable Energy}, vol. 224, p. 120152, 2024.
		
		\bibitem{b135} F. Fang, L. Tang, J. Nian, L. Wei and X. Wu, "A Few-Shot Wind Turbine Blade Surface Faults Detection via Enhanced Data Augmentation and Progressive Transfer Learning," \emph{IEEE Transactions on Instrumentation and Measurement}, vol. 75, pp. 1-15, 2026
		
		\bibitem{b14} X. Jia and X. Chen, ``AI-based optical-thermal video data fusion for near real-time blade segmentation in normal wind turbine operation,'' \emph{Engineering Applications of Artificial Intelligence}, vol. 127, p. 107325, 2024.
		
		\bibitem{b15} W. Zhou, Z. Wang, M. Zhang and L. Wang, ``Wind turbine actual defects detection based on visible and infrared image fusion,'' \emph{IEEE Transactions on Instrumentation and Measurement}, vol. 72, pp. 1--8, 2023.
		
		\bibitem{b16} P. Rizk, F. Rizk, S. S. Karganroudi, A. Ilinca, R. Younes and J. Khoder, ``Advanced wind turbine blade inspection with hyperspectral imaging and 3D convolutional neural networks for damage detection,'' \emph{Energy and AI}, vol. 16, p. 100366, 2024.
		
		\bibitem{b17} D. Dwivedi, K. V. S. M. Babu, P. K. Yemula, P. Chakraborty and M. Pal, ``Identification of surface defects on solar PV panels and wind turbine blades using attention based deep learning model,'' \emph{Engineering Applications of Artificial Intelligence}, vol. 131, p. 107836, 2024.
		
		\bibitem{b18} W. Li, W. Zhao and Y. Du, ``Large-scale wind turbine blade operational condition monitoring based on UAV and improved YOLOv5 deep learning model,'' \emph{Mechanical Systems and Signal Processing}, vol. 226, p. 112386, 2025.
		
		\bibitem{b19} X. Jia and X. Chen, ``Unsupervised wind turbine blade damage detection with memory-aided denoising reconstruction,'' \emph{IEEE Transactions on Industrial Informatics}, vol. 21, no. 1, pp. 762--770, 2025.
		
		\bibitem{b20} X. Lei, M. Sun, R. Zhao, H. Wu, Z. Zhou, Y. Dong and L. Sun, ``Unsupervised vision‐based structural anomaly detection and localization with reverse knowledge distillation,'' \emph{Structural Control and Health Monitoring}, vol. 2024, no. 1, p. 8933148, 2024.
		
		\bibitem{b21} A. Foster, O. Best, M. Gianni, A. Khan, K. Collins and S. Sharma, ``Drone footage wind turbine surface damage detection,'' in \emph{Proc. 2022 IEEE 14th Image, Video, and Multidimensional Signal Processing Workshop (IVMSP)}, pp. 1--5, 2022.
		
		\bibitem{b22} S. T. Ataei, P. M. Zadeh and S. Ataei, ``Vision-based autonomous structural damage detection using data-driven methods,'' \emph{arXiv preprint}, arXiv:2501.16662, 2025.
		
		\bibitem{b23} Y. Zhang, L. Wang, C. Huang and X. Luo, ``Wind turbine blade defect detection based on the genetic algorithm-enhanced YOLOv5 algorithm using synthetic data,'' \emph{IEEE Transactions on Industry Applications}, vol. 61, no. 1, pp. 653--665, 2025. DOI: 10.1109/TIA.2024.3481190.
		
		\bibitem{b24} X. Ye, L. Wang, C. Huang and X. Luo, ``Wind turbine blade defect detection with a semi-supervised deep learning framework,'' \emph{Engineering Applications of Artificial Intelligence}, vol. 136, p. 108908, 2024.
		
		\bibitem{b25} S. Sheiati, X. Jia, M. McGugan, K. Branner and X. Chen, ``Artificial intelligence-based blade identification in operational wind turbines through similarity analysis aided drone inspection,'' \emph{Engineering Applications of Artificial Intelligence}, vol. 137, p. 109234, 2024.
		
		\bibitem{b28} C. Yang, X. Liu, H. Zhou, Y. Ke and J. See, ``Towards accurate image stitching for drone-based wind turbine blade inspection,'' \emph{Renewable Energy}, vol. 203, pp. 267--279, 2023.
		
		\bibitem{b29} T. Li, Y. Luan, Z. Pang and W. Zhang, ``Structural digital twin modeling and adaptive pretrain-finetune learning for dynamic impact identification on wind turbine blades,'' \emph{IEEE Transactions on Industrial Informatics}, vol. 20, no. 8, pp. 10292--10303, 2024.
		
		\bibitem{b30} G. Jiang, R. Yue, Q. He, P. Xie and X. Li, ``Imbalanced learning for wind turbine blade icing detection via spatio-temporal attention model with a self-adaptive weight loss function,'' \emph{Expert Systems with Applications}, vol. 229, p. 120428, 2023.
		
		\bibitem{b31} Q. Zhou, G. Pang, Y. Tian, S. He and J. Chen, ``AnomalyClip: Object-agnostic prompt learning for zero-shot anomaly detection,'' \emph{arXiv preprint}, arXiv:2310.18961, 2023.
		
		\bibitem{b32} Z. Gu, B. Zhu, G. Zhu, Y. Chen, H. Li, M. Tang and J. Wang, ``FILO: Zero-shot anomaly detection by fine-grained description and high-quality localization,'' in \emph{Proc. 32nd ACM International Conference on Multimedia}, pp. 2041--2049, 2024.
		
		\bibitem{b33} M. H. Soleimani-Babakamali, R. Soleimani-Babakamali, K. Nasrollahzadeh, O. Avci, S. Kiranyaz and E. Taciroglu, ``Zero-shot transfer learning for structural health monitoring using generative adversarial networks and spectral mapping,'' \emph{Mechanical Systems and Signal Processing}, vol. 198, p. 110404, 2023.
		
		\bibitem{b34} M. H. Soleimani-Babakamali, R. Soleimani-Babakamali, A. Kashfi-Yeganeh, K. Nasrollahzadeh, O. Avci, S. Kiranyaz and E. Taciroglu, ``Multi-source transfer learning for zero-shot structural damage detection,'' \emph{Applied Soft Computing}, vol. 169, p. 112519, 2025.
		
		\bibitem{b35} Q. Xiong, Q. Kong, H. Xiong, J. Chen, C. Yuan, X. Wang and Y. Xia, ``Zero-shot knowledge transfer for seismic damage diagnosis through multi-channel 1D CNN integrated with autoencoder-based domain adaptation,'' \emph{Mechanical Systems and Signal Processing}, vol. 217, p. 111535, 2024.
		
		\bibitem{b36} M. Chen, S. Mangalathu and J. S. Jeon, ``Rapid damage state identification of structures using generalized zero‐shot learning method,'' \emph{Earthquake Engineering \& Structural Dynamics}, vol. 53, no. 14, pp. 4269--4286, 2024.
		
		\bibitem{b37} Z. Gu, B. Zhu, G. Zhu, Y. Chen, M. Tang and J. Wang, ``AnomalyGPT: Detecting industrial anomalies using large vision-language models,'' in \emph{Proc. AAAI Conference on Artificial Intelligence}, vol. 38, no. 3, pp. 1932--1940, 2024.
		
		\bibitem{b38} W. Chen, Y. Yan-yi, Z. Tie-zheng, P. Da-peng, T. Hui-han, L. Zhi, Y. Qing-wen, W. Hui-han and W. Ying-you, ``Systems engineering issues for industry applications of large language model,'' \emph{Applied Soft Computing}, vol. 151, p. 111165, 2024.
		
		\bibitem{b39} C. Walker, C. Rothon, K. Aslansefat, Y. Papadopoulos and N. Dethlefs, ``SafeLLM: Domain-specific safety monitoring for large language models: A case study of offshore wind maintenance,'' \emph{arXiv preprint}, arXiv:2410.10852, 2024.
		
		\bibitem{b40} S. Pastoriza, I. Yousfi, C. Redino, M. Vucovich, A. Rahman, S. Aguinaga, and D. Nandakumar, ``Retrieval Augmented Anomaly Detection (RAAD): Nimble Model Adjustment Without Retraining,'' in \emph{2025 13th International Symposium on Digital Forensics and Security (ISDFS)}, pp. 1--6, 2025.
		
		\bibitem{b41} H. Thimonier, F. Popineau, A. Rimmel and B. L. Doan, ``Retrieval augmented deep anomaly detection for tabular data,'' in \emph{Proc. 33rd ACM International Conference on Information and Knowledge Management}, pp. 2250--2259, 2024.
		
		\bibitem{b42} H. Wang and Y. F. Li, ``Large language model empowered by domain-specific knowledge base for industrial equipment operation and maintenance,'' in \emph{Proc. 2023 5th International Conference on System Reliability and Safety Engineering (SRSE)}, pp. 474--479, 2023.
		
		\bibitem{b425} P. LIU, L. Qian, X. Zhao and B. Tao, "Joint Knowledge Graph and Large Language Model for Fault Diagnosis and Its Application in Aviation Assembly," \emph{IEEE Transactions on Industrial Informatics}, vol. 20, no. 6, pp. 8160-8169, 2024.
	
		\bibitem{b43} C. Walker, C. Rothon, K. Aslansefat, Y. Papadopoulos and N. Dethlefs, ``Using large language models to recommend repair actions for offshore wind maintenance,'' in \emph{Journal of Physics: Conference Series}, vol. 2875, no. 1, p. 012025, 2024.
		
		\bibitem{b44} S. Alnegheimish, L. Nguyen, L. Berti-Equille and K. Veeramachaneni, ``Large language models can be zero-shot anomaly detectors for time series?'' \emph{arXiv preprint}, arXiv:2405.14755, 2024.
		
		\bibitem{b45} S. Jose, K. T. Nguyen, K. Medjaher, R. Zemouri, M. Lévesque and A. Tahan, ``Advancing multimodal diagnostics: Integrating industrial textual data and domain knowledge with large language models,'' \emph{Expert Systems with Applications}, vol. 255, p. 124603, 2024.
		
		\bibitem{b46} M. Bonomo and S. Bianco, ``Visual RAG: Expanding MLLM visual knowledge without fine-tuning,'' \emph{arXiv preprint}, arXiv:2501.10834, 2025.
		
		\bibitem{b47} N. N. Bhat, J. Mondal and S. Sarkar, ``ExpertNeurons at SciVQA-2025: Retrieval Augmented VQA with Vision Language Model (RAVQA-VLM),'' in \emph{Proc. Fifth Workshop on Scholarly Document Processing (SDP 2025)}, pp. 221--229, Jul. 2025.
		
		\bibitem{b48} N. Ueda, Y. Dong, K. Boros, D. Ito, T. Sera, and M. Oyamada, ``SCAN: Semantic Document Layout Analysis for Textual and Visual Retrieval-Augmented Generation,'' in \emph{Findings of the Association for Computational Linguistics: EACL 2026}, pp. 1618--1637, 2026.
		
		\bibitem{b49} F. F. Khan, J. Chen, Y. Mohamed, C. M. Feng and M. Elhoseiny, ``VR-RAG: Open-vocabulary Species Recognition with RAG-Assisted Large Multi-Modal Models,'' \emph{arXiv preprint}, arXiv:2505.05635, 2025.
		
		\bibitem{b50} X. Zheng, Z. Weng, Y. Lyu, L. Jiang, H. Xue, B. Ren, D. Paudel, N. Sebe, L. Van Gool and X. Hu, ``Retrieval augmented generation and understanding in vision: A survey and new outlook,'' \emph{arXiv preprint}, arXiv:2503.18016, 2025.
		
		\bibitem{b51} H. Chase, ``LangChain,'' 2022. [Online]. Available: \underline{https://github.com/langchain-ai/langchain}
		
		\bibitem{b52} N. Reimers and I. Gurevych, ``Sentence-BERT: Sentence Embeddings Using Siamese BERT-Networks,'' in \emph{Proceedings of the 2019 Conference on Empirical Methods in Natural Language Processing and the 9th International Joint Conference on Natural Language Processing (EMNLP-IJCNLP)}, pp.3982--3992, 2019.
		
		\bibitem{b53} A. Radford, J. W. Kim, C. Hallacy, A. Ramesh, G. Goh, S. Agarwal, G. Sastry, A. Askell, P. Mishkin, J. Clark and G. Krueger, ``Learning transferable visual models from natural language supervision,'' in \emph{Proc. International Conference on Machine Learning}, pp. 8748--8763, 2021.
		
		\bibitem{b54} M. Douze et al., "The Faiss Library,"  \emph{IEEE Transactions on Big Data}, vol. 12, no. 2, pp. 346-361, 2026.
		
		\bibitem{b55} J. Bai, S. Bai, S. Yang, S. Wang, S. Tan, P. Wang, J. Lin, C. Zhou and J. Zhou, ``Qwen-VL: A versatile vision-language model for understanding, localization, text reading, and beyond,'' \emph{arXiv preprint}, arXiv:2308.12966, 2023.
		
		\bibitem{b56} X. Chen, ``Drone-based optical and thermal videos of rotor blades taken in normal wind turbine operation,'' \emph{IEEE Dataport}, 2023. DOI: https://dx.doi.org/10.21227/yzs5-1067.
		
		\bibitem{b57} X. Chen, ``Dataset for AI-based optical-thermal video data fusion for near real-time blade segmentation in normal wind turbine operation,'' Mendeley Data, V1, 2024. DOI: 10.17632/9rcf5p89zn.1
		
		\bibitem{b58} A. Shihavuddin and X. Chen, ``DTU - Drone inspection images of wind turbine,'' 2018.
		
		\bibitem{b59} W. Wang, Y. Xue, C. He and Y. Zhao, ``Review of the typical damage and damage-detection methods of large wind turbine blades,'' \emph{Energies}, vol. 15, no. 15, p. 5672, 2022.
		
		\balance
		\bibitem{b60} Ultralytics, ``YOLOv8,'' GitHub repository, 2023. [Online]. Available: \underline{https://github.com/ultralytics/ultralytics}
		
	\end{thebibliography}
\end{document}